\documentclass{article}

\PassOptionsToPackage{numbers, compress}{natbib}

\usepackage[preprint]{neurips_2022}

\usepackage{amsmath}
\usepackage{graphicx}
\usepackage[utf8]{inputenc} 
\usepackage[T1]{fontenc}    
\usepackage{float}
\usepackage{hyperref}       
\usepackage{url}            
\usepackage{booktabs}       
\usepackage{amsfonts}       
\usepackage{nicefrac}       
\usepackage{microtype}      
\usepackage{xcolor}         

\usepackage{amsfonts}
\usepackage{tipa}
\usepackage{verbatim}
\usepackage{subfigure}
\usepackage{booktabs,makecell,multirow}

\usepackage{color}
\usepackage{bbm}
\usepackage[linesnumbered,ruled]{algorithm2e}
\usepackage{multicol}
\usepackage{wrapfig}
\usepackage{bm}
\usepackage{array}
\usepackage{authblk}

\title{Distributional Reward Estimation for Effective Multi-Agent Deep Reinforcement Learning}

\author{%
  $\text{\textbf{Jifeng Hu}}^{\bm{1}}$~~~~~~~~~~$\text{\textbf{Yanchao Sun}}^{\bm{2}}$~~~~~~~~~~$\text{\textbf{Hechang Chen}}^{\bm{3}}\thanks{Corresponding Author. Email address: chenhc@jlu.edu.cn (H. Chen) and lis221@lehigh.edu (L. Sun). }$~~~~~~~~~~$\text{\textbf{Sili Huang}}^{\bm{4}}$\\
  $\text{\textbf{Haiyin Piao}}^{\bm{5}}$~~~~~~~~~~$\text{\textbf{Yi Chang}}^{\bm{6}}$~~~~~~~~~~$\text{\textbf{Lichao Sun}}^{\bm{7}*}$\\
  ${}^{\bm{1,3,4,6}}$School of Artificial Intelligence, Jlilin University, Changchun, China\\
  ${}^{\bm{2}}$Department of Computer Science, University of Maryland, College Park, MD 20742, USA\\
  ${}^{\bm{5}}$Northwestern Polytechnical University, Xian, China\\
  ${}^{\bm{7}}$Lehigh University, Bethlehem, Pennsylvania, USA\\
  ${}^{\bm{1,4}}$\texttt{hujf21@mails.jlu.edu.cn, huangsl21@mails.jlu.edu.cn}\\
  ${}^{\bm{2}}$\texttt{ycs@umd.edu}\\
  ${}^{\bm{3,6}}$\texttt{chenhc@jlu.edu.cn, yichang@jlu.edu.cn}\\
  ${}^{\bm{5}}$\texttt{haiyinpiao@mail.nwpu.edu.cn}\\
  ${}^{\bm{7}}$\texttt{lis221@lehigh.edu}\\
}

\begin{document}

\maketitle


\begin{abstract}
  Multi-agent reinforcement learning has drawn increasing attention in practice, e.g., robotics and automatic driving, as it can explore optimal policies using samples generated by interacting with the environment. However, high reward uncertainty still remains a problem when we want to train a satisfactory model, because obtaining high-quality reward feedback is usually expensive and even infeasible. To handle this issue, previous methods mainly focus on passive reward correction. At the same time, recent active reward estimation methods have proven to be a recipe for reducing the effect of reward uncertainty. In this paper, we propose a novel Distributional Reward Estimation framework for effective Multi-Agent Reinforcement Learning (DRE-MARL). Our main idea is to design the multi-action-branch reward estimation and policy-weighted reward aggregation for stabilized training. Specifically, we design the multi-action-branch reward estimation to model reward distributions on all action branches. Then we utilize reward aggregation to obtain stable updating signals during training. Our intuition is that consideration of all possible consequences of actions could be useful for learning policies. The superiority of the DRE-MARL is demonstrated using benchmark multi-agent scenarios, compared with the SOTA baselines in terms of both effectiveness and robustness.
\end{abstract}

\section{Introduction}
Multi-agent reinforcement learning (MARL) has achieved substantial successes in solving real-time competitive games~\cite{openai2019dota4backup}, robotic manipulation~\cite{andrychowicz2020learning}, autonomous traffic control~\cite{palanisamy2020multi}, and quantitative trading strategies~\cite{lee2020maps}.  
Most existing works require agents to receive high-quality supervision signals, i.e., rewards, which are either infeasible or expensive to obtain in practice~\cite{wang2021policy}.
The rewards provided by the environment are subject to multiple kinds of randomness.
For example, the reward collected from sensors on a robot will be affected by physical conditions such as temperature and lighting, which makes the reward full of bias and intrinsic randomness.
The interplay among agents will result in more reward uncertainty. For instance, though an agent executes the same action under the same state, the reward can still vary because other agents can execute other actions.
Thus, handling non-stationary rewards in the learning process is a necessity for successfully learning complex behaviors in multi-agent environments.

In fact, there exist numerous works~\cite{majadas2021disturbing,natarajan2013learning,wang2021policy,wang2020reinforcement} in reinforcement learning (RL) that consider the \emph{reward uncertainty} during training.
The traditional stream of research on non-stationary feedback focuses on passive reward correction.
For example, \citet{wang2020reinforcement} adopt the confusion matrix to model the reward uncertainty and obtain a stable reward for learning. 
While in \cite{wang2021policy}, the authors recover the true supervision signals with peer loss, which punishes over-agreement for avoiding overfitting.
However, the assumption of the reward uncertainty of the above works is restrictive. 
For instance, the reward flipping mechanism~\cite{wang2020reinforcement} restricts the randomness of the reward to a countable value set.
Recently, another stream of research focuses on active reward estimation.
To alleviate the reward uncertainty issue, some works have treated reward estimation as a point-to-point regression problem where each state-action pair is mapped to a reward~\cite{romoff2018reward,zhang2020robust}.
However, the reward uncertainty, particularly caused by agents' interplay, can not be fully resolved by the point-to-point reward estimation in MARL.
This is because the regression is good at the one-to-one mapping between state and rewards.
But in MARL, the same state-action pair input of one agent will lead to multiple environmental rewards (i.e., one-to-many mapping\label{one-to-many mapping}), which is intractable for regression and thus hurts the performance.
Moreover, these methods do not consider the fact that the reward uncertainty comes not only from inherent environment randomness but also from the interplay among agents.
These two factors cause blended influence on the received reward and increase the training difficulty.
As we experimentally show later (See Figure~\ref{abalation} (left) for details.), such a point-to-point strategy leads to a worse suboptimal outcome with the increasing agent number and the reward uncertainty degree~\cite{ovadia2019can}.

In this work, we aim to develop distributional reward estimation followed by policy-weighted reward aggregation for MARL.
Intuitively, our idea is like one human constructs a reward blueprint of all action branches in his brain and thoughtfully makes a decision by considering all possible consequences.
Traditional methods~\cite{mnih2015human,schulman2017proximal} evaluate the policy only with environment rewards, which will bring about more uncertainty when training the critic because they only take into account the reward $r_k$ received after taking a specific action, i.e., $k$-th action $a_k$, from the policy.
While in our method, we consider not only the environmental rewards but also the potential rewards on other action branches to perform more stable critic updating and thus achieve better performance.  
So we propose multi-action-branch distributional reward estimation to model reward distributions $\{\tilde{R}^{i}(o^{i},a^{i}_{k})\}_{k=1}^{K}$ on all action branches, where $o^{i}_{t}$ is the observation and $a^{i}_{k,t}$ is the $k$-th action.
Then we aggregate the environment reward $r_k$ and rewards sampled from different action-branch distributions $\{\tilde{R}^{i}(o^{i},a^{i}_{m})\}_{m=1,m\neq k}^{K}$ by weighting them according to the corresponding action selection probability of the current policy.
We will obtain the aggregated mix reward $\bar{R}^{i}$ and lumped reward $\bar{r}^{i}$ for training each agent through the reward aggregation.
Reward aggregation enables the agent to evaluate its historical decisions thoughtfully, thus providing a sophisticated and effective way to reduce the influence of reward uncertainty.
Such a model covers the reward uncertainty in MARL environments and achieves better performance in several MARL benchmarks from small to large agent numbers.

Our contribution is three-fold.
1) We propose a novel framework, called Multi-agent Distributional Reward Estimation (DRE-MARL), to systematically characterize the reward uncertainty in MARL by modeling reward distributions for all action branches.
To the best of our knowledge, this is the first effort to solve the reward uncertainty of MARL with multi-action-branch distributional reward estimation followed by reward aggregation.
2) Policy-weighted reward aggregation is developed in our framework which enables us to perform stable training of the critic and actors.
Besides, DRE-MARL is a universal framework that can be expediently integrated into other MARL methods.
3) We validate the performance of our algorithm with function approximation and mini-batch update via extensive simulations in MARL scenarios with different agent numbers and reward uncertainty.

\section{Related Work}
\noindent\textbf{Reward Uncertainty.}~~~
Dealing appropriately with reward uncertainty has received quite a bit of attention in recent reinforcement learning studies~\cite{wang2021policy,wang2020reinforcement,neider2022reinforcement,lee2020weakly,sukhbaatar2014learning,menon2015learning,li2021provably,natarajan2013learning,scott2015rate,van2015learning}.  
The core idea of this line of works is to assume access to the knowledge of the noise and define unbiased surrogate loss functions to recover the true loss or reward.  
A typical seminal work dates back to~\cite{natarajan2013learning}, which recovers the true loss from the noisy label distribution with the knowledge of the noise rates of labels.
Follow-up works offer solutions to estimate the uncertainty level from model predictions~\cite{blanchard2016classification,zhu2021second,zhu2021federated,sun2020vulnerability,sukhbaatar2014learning} or clusterable representations~\cite{zhu2021clusterability}.
A couple of recent works~\cite{wang2020reinforcement,zhu2021federated,neider2022reinforcement} have also looked into this problem in sequential settings.   
For example, \citet{everitt2017reinforcement} analyze the potential sources of uncertainty and provide the impossibility result for training under facultative reward uncertainty.  
\citet{wang2020reinforcement} consider modeling the reward uncertainty being caused by a confusion matrix and design a statistics-based estimation method to cover the uncertainty. 
\citet{wang2021policy} recover the true supervision signals with peer loss, which punishes over-agreement to avoid overfitting.
But the method is inefficient in MARL, because the uncertainty is bigger than single agent RL.

\textbf{Reward Estimation.}~~~~A number of prior literature in robotic and non-robotic domains have adopted reward estimation~\cite{franccois2019combined,romoff2018reward,feinberg2018model,silver2017predictron,henaff2017model,sun2022transfer,racaniere2017imagination,sutton1990integrated,grzes2009learning,sun2017collaborative,sorg2012variance,talvitie2018learning,sun2019can,metelli2017compatible}.  
In most of these cases, the predicted reward is used for planning. 
They take advantage of the estimated reward in imagination augmented rollouts with a dynamic model (i.e., world model) accompanying the reward estimator.
However, in our case, we aim at handling reward uncertainty rather than planning with multi-action-branch reward estimation and aggregation, thus avoiding learning system dynamics and multi-step imaginary rollouts.  
Reward estimation can also be performed by reward shaping~\cite{yuan2021multimodal,hlynsson2021reward} (RS) and inverse reinforcement learning~\cite{hadfield2017inverse,liang2021reward} (IRL). But IRL doesn't account for the reward uncertainty, and RS focuses on efficient exploration.
Recently, \citet{romoff2018reward} proposes to train reward estimator with function approximation alongside the value function. However, it adopts point-to-point reward estimation, which struggles when generalizing to multi-agent settings.  

\textbf{Distributional RL.}~~~~Distributional RL has recently gained increasing attention due to its powerful capacity for treating RL's uncertainty~\cite{sun2021dfac,sun2021distributional,qiu2021rmix,rockafellar2000optimization,bellemare2017distributional,rowland2019statistics,philippe2001value,dabney2018distributional}. Early studies about return uncertainty can be dated back to \citet{sobel1982variance}. In MARL, the uncertainty issue is a more challenging problem than single RL. Fortunately, many studies focus on it~\cite{sun2021dfac,sun2021distributional,qiu2021rmix}. For example, \citet{sun2021dfac} propose the Mean-Shape Decomposition method and quantile mixture in value decomposition, bridging the gap between distributional RL and value function factorization. Although our work also belongs to the distributional field, we focus on the reward uncertainty of the environment rather than Q-value uncertainty and model the distribution of reward rather than Q compared with \cite{sun2021dfac,sun2021distributional,qiu2021rmix}. Besides, we propose reward estimation followed by reward aggregation, which is novel and effective.

\vspace{-0.2em}
\section{Preliminaries}

A Markov game~\cite{littman1994markov,sun2021temple}, a multi-agent extension of the Markov decision process, with partial observability can be described as the tuple $\mathcal{M}=\langle\mathcal{N}, \mathcal{S}, \{\mathcal{A}^{i}\}_{i\in \mathcal{N}}, \{\mathcal{O}^{i}\}_{i\in \mathcal{N}}, \mathcal{P}, \{R^i\}_{i\in \mathcal{N}}, \gamma\rangle$, where $\mathcal{N}=\{1,...,N\}$ is the set of $N$ agents, $\mathcal{S}$ is the state space of the environment, $\mathcal{A}^{i}$ is the action space of agent $i$, $\mathcal{O}^{i}$ is the observation space of agent $i$, $\mathcal{P}: \mathcal{S} \times \mathcal{A}^{1}\times \cdot\cdot\cdot\times\mathcal{A}^{N} \rightarrow \Delta(\mathcal{S})$ ($\Delta(x)$ represents the space of distributions over $x$.) denotes the state transition probability that is a mapping from the current state and the joint action to the probability distribution over the state space, and $\gamma\in[0,1)$ is the discounting factor. We consider the fact that the interplay between agents results in distributional reward feedback: instead of observing $R^{i}$ for each agent $i$, agent $i$ can only observe the reward sampled from the reward distribution $\widehat{R}^i:\mathcal{S}\times\mathcal{A}^{1}\times \cdot\cdot\cdot\times\mathcal{A}^{N}\rightarrow\Delta(\mathbb{R})$.

At each time t, $N$ agents receive different observations $(o^{1}_{t},...,o^{N}_{t})$ and output certain joint action $\boldsymbol{a_{t}}=\{a^{i}_{k,t}\}_{i\in\mathcal{N}}, k\in\{1,...,K\}$, where $a^{i}$ is executed according to agent $i$'s policy $\pi^i:\mathcal{O}^i\rightarrow\Delta(\mathcal{A}^i)$, $K=|\mathcal{A}|$ is the size of action space. 
The environment then transitions to $s_{t+1}$ and rewards each agent $i$ by $r^{i}_{t}=R^{i}(s_t, \boldsymbol{a_{t}}, s_{t+1})$, where $R^{i}(\cdot)$ is the reward function. 
The goal of agent $i$ is to find the best policy $\pi^{i}$ that maximizes the total reward received from the environment from a start state to an end state. The expected cumulative discounted reward is expressed as $\mathbb{E}[\sum_{t=0}^\infty\gamma^{t}\cdot r^{i}_{t}]$.
In this paper, we consider the $N$ agents, represented by $\boldsymbol{\pi}=\{\pi^{1},...,\pi^N\}$, with policies (actors) parameterized by $\boldsymbol{\theta}=\{\theta_1, ..., \theta_N\}$. 
Given a specific policy $\pi$, we adopt state value function the $V^{\pi}_{\gamma,\psi}$ to approximate expected return, where $\psi$ is the parameters of $V^{\pi}_{\gamma,\psi}$.

\vspace{-0.2em}
\section{Problem Formulation}

\subsection{Reward Uncertainty in MARL}

Reward uncertainty is prevalent in many real-world scenarios, but its influence is not well solved in the literature. Handling non-stationary reward during training is necessary for successfully learning complex behaviors in multi-agent environments.
In many practical scenarios, the agents usually observe non-stationary reward feedback due to the interplay and inherent randomness rather than perfectly receiving the precise reward for training.
Consider a multi-agent environment with high complexity and large state-action space such that it is intractable for agents to explore the entire state space. Suppose that we can only have access to the non-stationary reward.
How can the agents learn cooperative behaviors under the reward uncertainty?
We should figure out the factors that cause the reward uncertainty. Assume we are given a multi-agent scenario consisting of $N$ agents.
the reward contains uncertainty that is caused by multiple factors that can be classified into two aspects.

\textbf{Mutual interaction.}\label{Mutual interaction}~~The first factor that causes reward uncertainty is mutual interaction between agents. As mentioned in Section~\ref{one-to-many mapping}, the interplay between agents can lead to the one-to-many mapping between state and reward. For example, suppose that we need two agents to complete a task cooperatively.
The agent will receive various rewards because its partner executes other actions even though the agent executes the same action under the same observation.

\textbf{Natural disturbance.}~~ Another aspect lies in the natural disturbance of the environment. During the training process, the environment contains inherent randomness that is ubiquitous in the real world. For instance, the sensor's feedback fluctuates due to the variation in temperature and lighting, which makes the reward feedback inaccurate. 

In the following section, we focus on the two aspects and propose multi-action-branch reward estimation and reward aggregation for providing stable updating signals for agents. Here, the intuition is that the decision should be executed after carefully considering all possible consequences.
Finally, in Section~\ref{exp}, we empirically investigate the validity of our method in different scenarios.

\subsection{Reward Estimation}
Usually, many MARL scenarios provide precise human-designed rewards, which deviate from the actual situation in practice.
To simulate the reward uncertainty, we define the stochastic rewards that are provided by the environment according to some stochastic processes.
That is, the reward is generated from some distributions with certain probability densities~\cite{wang2020reinforcement}.
To capture the reward uncertainty and facilitate training, reward estimation is proposed to reduce the impact of stochastic rewards~\cite{wang2020reinforcement, romoff2018reward}.
The reward estimator~\cite{rudner2021outcome}, which is a core component in a scenario with stochastic rewards, evaluates $s$, $(s,a)$, or $(s,a,s^{\prime})$ pairs at time step $t$ and estimates the possible rewards that are used for guiding the agent through the user goal. 
\citet{romoff2018reward} model the task of learning the reward estimator as a simple point-to-point regression problem by function approximation: $\mathcal{L}(\varphi)=\mathbb{E}[(r-\tilde{R}(\mathcal{T};\varphi))^{2}]$, where $\tilde{R}(\mathcal{T};\varphi)$ is the reward estimator with parameter $\varphi$ based on different inputs $\mathcal{T}\in\{s, (s,a), (s,a,s^{\prime})\}$.

\begin{figure*}[t!]
\vspace{-1em}
 \begin{center}
 \includegraphics[angle=0,width=0.93\textwidth]{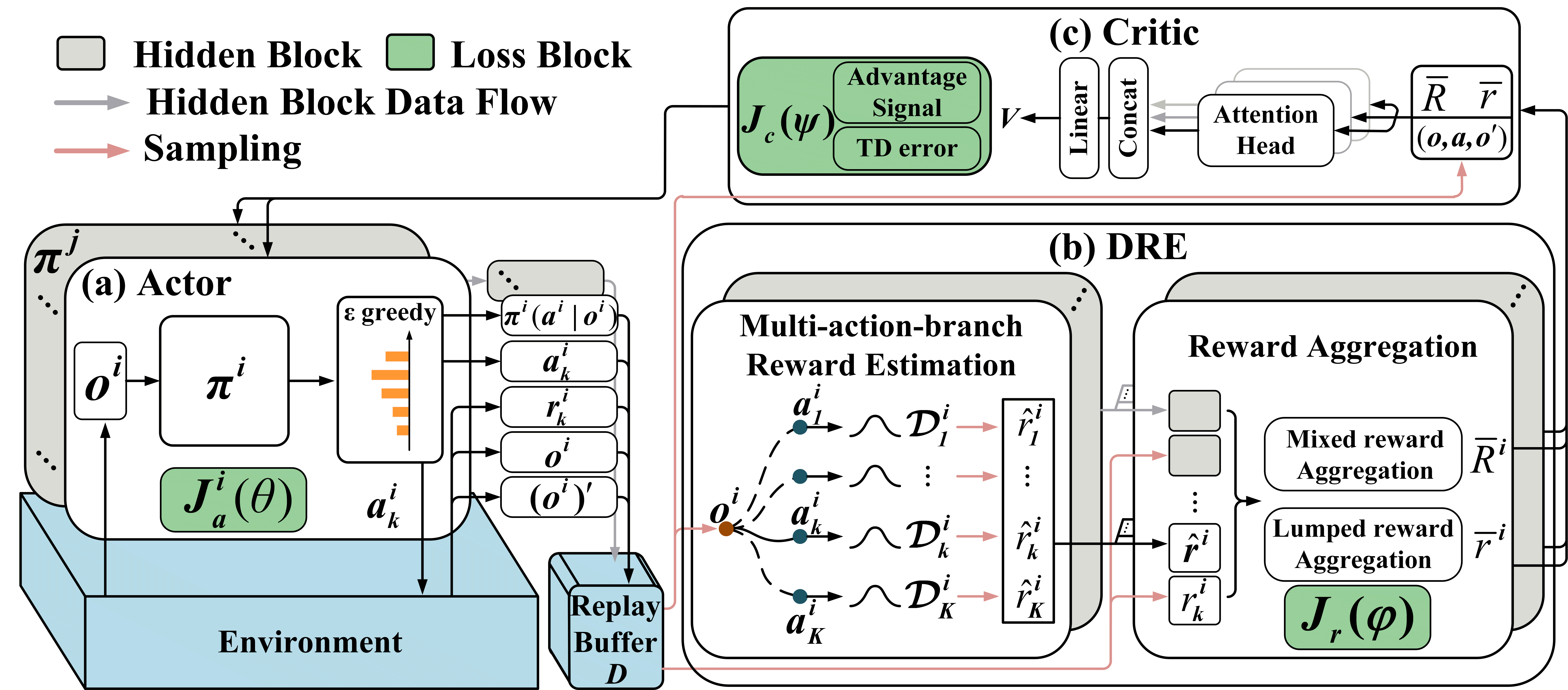}
 \caption{The overall architecture of Multi-agent Distributional Reward Estimation, which contains \textbf{(a) Decentralized Actors}~\textbf{(b) Distributional Reward Estimation}~\textbf{(c) Centralized Critic}. As is shown in \textbf{(a)}, we perform decentralized execution according to each agent's observation and store experience in the replay buffer. Then in \textbf{(b)}, we perform multi-action-branch reward estimation followed by policy-weighted reward aggregation. Finally, as shown in \textbf{(c)}, the centralized critic executes training with aggregated rewards and provides advantage signals for actors.}
 \label{framework}
 \end{center}
 \end{figure*}

\vspace{-0.1mm}
\section{DRE-MARL}
In this section, we introduce a general training framework, Multi-agent Distributional Reward Estimation (DRE-MARL), for the reward uncertainty problem in MARL, as shown in Figure~\ref{framework}.
We adopt the architecture of centralized training and decentralized execution (CTDE)~\cite{lowe2017multi}, which consists of $N$ decentralized actors $\{\pi^{i}_{\theta}\}_{i\in\mathcal{N}}$ and a centralized critic $V_{\gamma,\psi}^{\pi}$, parameterized by $\theta$ and $\psi$, respectively.
In practice, we utilize $\{\pi^{i}_{\theta}\}_{i\in\mathcal{N}}$ shown in Figure~\ref{framework}(a) to interact with the environment and collect experiences.
Figure~\ref{framework}(b) shows the proposed distributional reward estimation structure, which consists of two stages: 
we first perform multi-action-branch reward distribution estimation from observed experiences and sample rewards from the reward distributions of action branches.
Then we aggregate the environment rewards and sampled rewards to guide the training of the critic and actors.
Figure~\ref{framework}(c) indicates that we utilize a graph attention network with multi-head attention~\cite{liu2020multi,chen2020gama} to capture the global information from observations $(o^{1},...,o^{N})$. 
The centralized critic can simultaneously produce advantage signals for all agents through one forward calculation during training.
For stability, we construct target centralized critic $V_{\gamma,\tilde{\psi}}$\cite{mnih2015human,silver2014deterministic} and $N$ target policies $\tilde{\pi}_{\tilde{\theta}}$, parameterized by $\tilde{\psi}$ and $\tilde{\theta}$, respectively. 
The target network parameters $(\tilde{\psi},\tilde{\theta})$ are only updated with current network parameters $(\psi,\theta)$ and are held fixed between individual updates~\cite{lillicrap2015continuous,sun2021strongest}.

\subsection{Multi-action-branch Reward Estimation}\label{Multi-action-branch Reward Estimation}

Modeling reward distributions is challenging because the reward uncertainty caused by agents' interplay grows exponentially with an increase of agent number. 
One direct method is to model reward distributions on joint state-action space. But this method suffers from huge uncertainty caused by mutual interaction as agent number grows.
Besides, in the ablation study (See Figure~\ref{abalation} (left) for details.), we also verify that modeling joint reward distributions cannot resolve the issue.

To handle the above challenge, we propose another method to achieve our goal. 
We simplify the issue by viewing other agents as a part of the environment such that we only need to focus on the reward estimation of each agent instead of all agents.
Additionally, inspired by the fact that humans will imagine the potential consequences based on historical experience, we equip each agent with a reward estimator $\widehat{R}^{i}$ to better capture the reward uncertainty and stabilize the training process.
Specifically, for agent $i$ at time step $t$, we propose the multi-action-branch reward estimator $\widehat{R}^{i}(o^{i}_{t},a^{i}_{k,t};\varphi^{i})\in\mathcal{D}$ to model the reward distribution based on agent $i$'s observation $o^{i}_{t}$ and $k$-th action $a^{i}_{k,t}$, where $\mathcal{D}$ represents reward distribution space, $\widehat{R}^{i}_{k}$ represents the reward distribution of agent $i$ in $k$-th action branch, and $\varphi^{i}$ is the parameters of $\widehat{R}^{i}$. We use $\widehat{R}^{i}\in{\mathcal{D}^{K}}$ to represent the $K$ estimated reward distributions for agent $i$. 
We can achieve the above goal by optimizing the overall objective function $J_{r} = \sum_{i}J_{r}^{i}(\varphi^{i})$, and the loss function of every agent is as follows: 
\begin{equation}\label{reward update}
    \begin{aligned}
        J_{r}(\varphi)=\mathbb{E}_{(o,a_{k},r_{k})\sim D}&\left[-\text{log}~\mathbb{P}[r_{k}|\widehat{R}(o,a_{k};\varphi)]+\mathcal{L}_{\widehat{R}}\right],
    \end{aligned}
\end{equation}
where we omit the superscript $i$ of $J_{r}(\varphi)$, $D$ is the replay buffer, and $-\text{log}~\mathbb{P}[\cdot|\cdot]$ is negative log likelihood. $\mathcal{L}_{\widehat{R}}$ is a regular term of reward distributions. 
In practice, we adopt Gaussian reward distribution $\mathcal{D}(\bm{\mu}, \bm{\sigma})$, but $\widehat{R}$ can be easily extended to other distributions. 
Then $\mathcal{L}_{\widehat{R}}$ can be defined by $\alpha\cdot\Vert\bm{\sigma}\Vert_{1}+\beta\cdot var(\bm{\mu})$, where $var$ is the variance of $\bm{\mu}$. $\alpha$ and $\beta$ are hyperparameters.

Multi-action-branch reward estimation enables us to forecast possible rewards of all action branches. 
It is just like we only think about each specific situation separately, but in practice, we usually make a decision considering all possible consequences.
Therefore, to evaluate the experience thoughtfully and stabilize training, we propose reward aggregation introduced in the following section to reduce the impact of reward uncertainty during the training process.

\subsection{Training with Reward Aggregation}\label{aggregation}
For each agent at each time step, the agent can only obtain a single reward $r_k$ while executing $k$-th action $a_k$.
But we augment $r_k$ as a built-up reward vector, where the $k$-th of the vector is $r_k$ and the following parts are the estimated rewards $\bm{\hat{r}}$ sampled from $\widehat{R}^{i}$.
Then we aggregate the built-up reward vector with the policy-weighted operation. 
We mainly obtain two types of rewards after reward aggregation: the mixed reward $\Bar{R}\in\mathbb{R}$ and the lumped reward $\bar{r}\in\mathbb{R}$, where $\bar{R}$ is used to update centralized critic $V_{\gamma,\psi}$ and $\bar{r}$ is used to update decentralized actors $\{\pi^{i}_{\theta}\}_{i\in\mathcal{N}}$.

Specifically, for each agent we first define the built-up reward vector $m=h(\bm{\hat{r}},r_k)\in\mathbb{R}^{K}$, which is constructed by replacing the $k$-th value of vector $\bm{\hat{r}}$ with true environment reward $r_k$ because the agent only takes $a_k$ during interaction with the environment, where $h(\cdot)$ is defined by $h(\bm{\hat{r}},r_k)=[\hat{r}_{1},...,r_{k},...,\hat{r}_{K}]$.
Then the mixed reward $\bar{R}=g(m^{1},...,m^{N})\cdot\Tilde{\pi}^{i}(\cdot|o)$ is calculated by policy-weighted aggregation, where $g(\cdot)$ represents two operations as below. 
1) \emph{MeanOperation} ($g_{MO}$): we utilize the average value of $m^{1},...,m^{N}$ as the output of $g(\cdot)$.  
2) \emph{SimpleSelection} ($g_{SS}$): we directly select $m^{i}$ for agent $i$ to calculate the mixed reward $\bar{R}$. Mathematically, $\bar{R}$ is defined by
\begin{equation}\label{reward aggregation R}
\bar{R}^{i}=g(m^{1},...,m^{N})\cdot\Tilde{\pi}^{i}(\cdot|o)=
    \begin{cases}
        mean(m^{1},...,m^{N})\cdot\Tilde{\pi}^{i}(\cdot|o) &\text{if } g=g_{MO}\\
        m^{i}\cdot\Tilde{\pi}^{i}(\cdot|o)  &\text{if } g=g_{SS}
    \end{cases}.
\end{equation}

Sampling from the reward distributions has been proven to be beneficial for exploration in some settings~\cite{liang2021reward,wang2020reinforcement,majadas2021disturbing}. 
But at the same time, it will also introduce more uncertainty during training. 
Here, we mainly focus on reducing the influence of reward uncertainty. Therefore we use the mean value of reward distributions to perform reward aggregation, usually achieving a more stable training process.
These two schemes, the sampling option and mean option, can be chosen flexibly. 
It is a tradeoff: we choose the former option while the environment is hard to explore. The latter is more suitable for an environment with more uncertainty but is easy to explore.  

Next, the centralized critic updates a parametric state value function $V_{\gamma, \psi}$ by minimizing the Bellman residual
\begin{equation}
    \begin{aligned}
        J_{c}(\psi)=\mathbb{E}_{(o,r_k,o^{\prime})\sim D}\left[\bar{R}+\gamma V_{\gamma,\tilde{\psi}}(o^{\prime})-V_{\gamma,\psi}(o)
        \right]^{2},
    \end{aligned}
\end{equation}
where $\psi$ and $\Tilde{\psi}$ are the parameters of current state value network and target state value network, respectively.
Finally, for each agent during each training iteration, the decentralized actors follow the same training scheme of policy optimization with advantage function~\cite{schulman2017proximal}:
\begin{equation}
    \begin{aligned}
        J_{a}(\theta)=\mathbb{E}\left[\text{min}\left(u\left(\theta\right), \text{clip}\left(u\left(\theta\right), 1-\epsilon, 1+\epsilon\right)\right)\hat{A}_{\gamma}+{\eta}H(\pi_{\theta})\right],
    \end{aligned}
\end{equation}
where $u(\theta)=\frac{\pi_{\theta}(a|o)}{\pi_{\tilde{\theta}}(a|o)}$ is the importance weight, $\epsilon$ is clip hyperparameter and usually we choose $\epsilon=0.2$. 
$H(\pi^{i}_{\theta})$ is the policy entropy of agent $i$, and $\eta$ controls the importance of entropy penalty objective.
For agent $i$, the advantage value is defined as follows: $\hat{A}^{i}_{\gamma}=\bar{r}^{i}+\gamma V_{\gamma,\tilde{\psi}}(o^{\prime})-V_{\gamma,\psi}(o)$, where $\bar{r}^{i}=l(m^{1},...,m^{N}, r_k)$ is the lumped reward that is defined by
\begin{equation}\label{reward aggregation r}
    \bar{r}^i=l(m^{1},...,m^{N}, r^{i}_k)=
        \begin{cases}
            mean(m^{1},...,m^{N}) &\text{if } l=l_{MO}\\
            mean(m^{i}) &\text{if } l=l_{SMO}\\
            r^{i}_k &\text{if } l=l_{SS}
    \end{cases},
\end{equation}
where $l$ represents three operations as follows. 1) \emph{MeanOperation} ($l_{MO}$): we average on all agents' built-up rewards ${m^{i}}_{i\in\mathcal{N}}$ to obtain a single output.
2) \emph{SimpleMeanOperation} ($l_{SMO}$): we utilize the mean value of agent $i$'s built-up rewards $m^{i}$ as output.
3) \emph{SimpleSelection} ($l_{SS}$): we directly select the environment reward $r^{i}_k$ as the lumped reward.

Following the above procedure, we can update the reward estimators and policies iteratively.
Further details and the full algorithm for optimizing DRE-MARL can be found in Appendix~\ref{algorithm}.

\section{Experiments} \label{exp}
\subsection{Environment Settings}
To demonstrate the effectiveness of the proposed method, we provide experimental results in several benchmark MARL environments based on the multi-agent particle environments~\cite{lowe2017multi} (MPE) and several variants of MPE.
Specifically, we consider cooperative navigation (CN), reference (REF), and treasure collection (TREA) environments from small to large agent numbers. 
In the following sections, we use CN-$q$, REF-$q$, and TREA-$q$ to represent different environment variants, where $q\in\{2,3,7,10\}$ and the entire environment variants are \{$\text{CN-}3$, $\text{CN-}7$, $\text{CN-}10$, $\text{REF-}2$, $\text{REF-}7$, $\text{REF-}10$, $\text{TREA-}3$, $\text{TREA-}7$, $\text{TREA-}10$\}.
In all these environments, the reward is set to collaborative, which means the reward is the summation of individual rewards. 
Detailed experimental description and settings can be found in Appendix~\ref{Environmental description}.

\noindent\textbf{Reward formulation.}~~~~
To simulate the reward uncertainty in these classic environments, we investigate several reward settings with induced uncertainty as follows. 1) \emph{Deterministic reward} ($r_{dete}$): in classic environments, the reward is calculated with respect to distance~\cite{iqbal2019actor,lowe2017multi}. Suppose that the feedback is exactly precise. The deterministic reward refers to that the same observation-action pair is mapped to an identical reward. We use $r^{i}_{dete}$ to represent this reward setting. 
2) \emph{Natural disturbance reward} ($r_{dist}$): reward uncertainty caused by natural disturbance, mentioned in Section~\ref{Mutual interaction}, can be regarded as the overall impact on the received reward because the sources of dynamics are coupled with each other. 
Similar to prior works~\cite{romoff2018reward,wang2020reinforcement}, we consider the natural disturbance reward that is designed by $r^{i}_{dist}\sim \mathcal{N}(r^{i}_{dete},1)*0.05+r^{i}_{dete}$.
We use $r^{i}_{dist}$ to represent this reward setting.
3) \emph{Mutual interaction reward} ($r_{ac-dist}$): in order to simulate the reward uncertainty caused by mutual interaction mentioned in Section~\ref{Mutual interaction}, we add different reward distributions on different action branches such that the same observation-action pair will result in different rewards. 
Specifically, the action dist-reward is defined by $r^{i}_{ac-dist}\sim\mathcal{N}(u, \delta)+r^{i}_{dete}$, where $u=\text{Index}(a^{i}_{k})$ and $\delta=0.001$.
We use $r^{i}_{ac-dist}$ to represent this reward setting.
$r_{dist}$ and $r_{ac-dist}$ settings are particularly challenging because agents need to conquer more dynamics.

\begin{table*}[t!]
\centering
\small
\caption{Performance comparison of DRE-MARL and several SOTA MARL algorithms under the $r_{dete}$, $r_{dist}$, and $r_{ac-dist}$ settings. The values represent mean episodic rewards.}
\resizebox{\textwidth}{!}{
\begin{tabular}{c|p{1.1cm}<{\raggedright} p{0.3cm}<{\raggedleft}|r r r r r r r}
\toprule
\specialrule{0em}{1.5pt}{1.5pt}
\toprule
 & Scenario & $q$ & 
 \makecell[c]{DRE-MARL\\(\textbf{ours})} & \makecell[c]{p2p-MARL\\\cite{romoff2018reward}} & \makecell[c]{MAPPO\\\cite{yu2021surprising}} & \makecell[c]{MAAC\\\cite{iqbal2019actor}} &  \makecell[c]{QMIX\\\cite{rashid2018qmix}} & \makecell[c]{MADDPG\\\cite{lowe2017multi}} & \makecell[c]{IQL\\\cite{tan1993multi}}\\
\toprule
\multirow{9}{*}{\rotatebox{90}{$r_{dete}$}} &   &3 & \textbf{-154\tiny{$\pm$11.90}} & -167\tiny{$\pm$12.80} & -425\tiny{$\pm$99.47} & -227\tiny{$\pm$43.38} & -1168\tiny{$\pm$77.73} & -169\tiny{$\pm$6.257} & -738\tiny{$\pm$175.2}\\
& CN-$q$ &7 & \textbf{-3070\tiny{$\pm$89.18}} & -3101\tiny{$\pm$121.5} &   -5674\tiny{$\pm$801.5} & -3293\tiny{$\pm$279.4} & -9320\tiny{$\pm$1528.} & -3428\tiny{$\pm$45.11} & -5092\tiny{$\pm$509.5}\\
&   &10 & \textbf{-8138\tiny{$\pm$285.6}}  & -8418\tiny{$\pm$277.0} & -13375\tiny{$\pm$1372.} & -8555\tiny{$\pm$873.0} & -21858\tiny{$\pm$4320.} & -8938\tiny{$\pm$67.94} & -12658\tiny{$\pm$864.4}\\
\cline{2-10}
&   &2 & -54\tiny{$\pm$8.589}  & -58\tiny{$\pm$8.638} & -85\tiny{$\pm$39.74} & -69\tiny{$\pm$28.30} & -438\tiny{$\pm$205.0} & \textbf{-44\tiny{$\pm$2.394}} & -95\tiny{$\pm$17.50}\\
& REF-$q$ &7 & \textbf{-732\tiny{$\pm$73.85}}  & -808\tiny{$\pm$88.99} & -1187\tiny{$\pm$365.0} & -1120\tiny{$\pm$302.5} & -1359\tiny{$\pm$130.5} & -1133\tiny{$\pm$50.92} & -2400\tiny{$\pm$285.2}\\
&   &10 & \textbf{-1524\tiny{$\pm$140.9}}  & -1542\tiny{$\pm$152.4} & -3317\tiny{$\pm$882.3} & -2380\tiny{$\pm$604.9} & -4037\tiny{$\pm$235.3} & -3102\tiny{$\pm$135.1} & -4921\tiny{$\pm$449.2}\\
\cline{2-10}
&   &3 & -458\tiny{$\pm$26.73}  & -457\tiny{$\pm$28.06} & -671\tiny{$\pm$95.23} & -480\tiny{$\pm$52.52} & \textbf{-390\tiny{$\pm$37.96}} & -428\tiny{$\pm$11.17} & -924\tiny{$\pm$86.37}\\
& TREA-$q$ &7 & \textbf{-2641\tiny{$\pm$134.6}}  & \textbf{-2638\tiny{$\pm$134.8}} & -3648\tiny{$\pm$0.005} & -2809\tiny{$\pm$270.0} & -3324\tiny{$\pm$270.9} & -2678\tiny{$\pm$59.59} & -4733\tiny{$\pm$246.7}\\
&   &10 & \textbf{-5076\tiny{$\pm$190.4}}  & -5260\tiny{$\pm$280.8} & -6864\tiny{$\pm$825.7} & \textbf{-4931\tiny{$\pm$438.1}} & -8715\tiny{$\pm$265.4} & -5699\tiny{$\pm$150.8} & -8465\tiny{$\pm$399.8}\\
 \midrule[1pt]
 \multirow{9}{*}{\rotatebox{90}{$r_{dist}$}} &   &3 & \textbf{-161\tiny{$\pm$13.74}}  & -186\tiny{$\pm$13.73} & -773\tiny{$\pm$103.9} & -270\tiny{$\pm$55.01} & -1263\tiny{$\pm$66.95} & -213\tiny{$\pm$6.300} & -552\tiny{$\pm$58.18}\\
& CN-$q$ &7 & \textbf{-3200\tiny{$\pm$93.43}}  & -3262\tiny{$\pm$132.3} & -7481\tiny{$\pm$1072.} & -3558\tiny{$\pm$419.7} & -10678\tiny{$\pm$2076.} & -3617\tiny{$\pm$49.06} & -6141\tiny{$\pm$1161.}\\
&  &10 & \textbf{-8605\tiny{$\pm$277.0}}  & -8792\tiny{$\pm$302.9} & -19696\tiny{$\pm$2607.} & -8986\tiny{$\pm$902.7} & -22699\tiny{$\pm$4151.} & -9365\tiny{$\pm$108.5} & -12693\tiny{$\pm$2046.}\\
\cline{2-10}
&   &2 & -57\tiny{$\pm$8.504}  & -63\tiny{$\pm$8.953} & -86\tiny{$\pm$27.21} & -72\tiny{$\pm$30.56} & -280\tiny{$\pm$193.1} & \textbf{-43\tiny{$\pm$2.234}} & -226\tiny{$\pm$30.06}\\
& REF-$q$ &7 & \textbf{-752\tiny{$\pm$79.95}} & -800\tiny{$\pm$88.33} & -1947\tiny{$\pm$639.3} & -1070\tiny{$\pm$283.7} & -1894\tiny{$\pm$195.3} & -1182\tiny{$\pm$38.16} & -2514\tiny{$\pm$283.6}\\
&   &10 & \textbf{-1676\tiny{$\pm$177.6}}  & \textbf{-1570\tiny{$\pm$156.0}} & -5580\tiny{$\pm$1745.} & -2558\tiny{$\pm$588.7} & -3584\tiny{$\pm$338.6} & -2968\tiny{$\pm$121.0} & -4090\tiny{$\pm$1140.}\\
\cline{2-10}
&   &3 & -484\tiny{$\pm$25.97}  & -447\tiny{$\pm$26.12} & -691\tiny{$\pm$136.3} & -513\tiny{$\pm$76.72} & \textbf{-427\tiny{$\pm$39.78}} & -449\tiny{$\pm$8.016} & -889\tiny{$\pm$49.29}\\
& TREA-$q$ &7 & \textbf{-2733\tiny{$\pm$131.6}}  & \textbf{-2745\tiny{$\pm$150.0}} & -3741\tiny{$\pm$0.000} & -2956\tiny{$\pm$313.2} & -3553\tiny{$\pm$212.7} & -2824\tiny{$\pm$75.14} & -5092\tiny{$\pm$394.1}\\
&   &10 & \textbf{-5340\tiny{$\pm$223.6}}  & -5515\tiny{$\pm$305.1} & -6912\tiny{$\pm$876.3} & \textbf{-5306\tiny{$\pm$587.2}} & -8897\tiny{$\pm$285.3} & -5847\tiny{$\pm$147.4} & -8727\tiny{$\pm$464.4}\\
\midrule[1pt]
 \multirow{9}{*}{\rotatebox{90}{$r_{ac-dist}$}} &   &3 & \textbf{477\tiny{$\pm$59.64}}  & 271\tiny{$\pm$87.00} & 156\tiny{$\pm$80.78} & -98\tiny{$\pm$254.8} & 367\tiny{$\pm$61.66} & 476\tiny{$\pm$67.32} & -2645\tiny{$\pm$740.2}\\
& CN-$q$ &7 & \textbf{888\tiny{$\pm$398.3}} & -480\tiny{$\pm$484.0} & -2960\tiny{$\pm$1418.} & -85\tiny{$\pm$487.5} & -7122\tiny{$\pm$1167.} & 802\tiny{$\pm$345.6} & -24492\tiny{$\pm$4691.}\\
&   &10 & \textbf{133\tiny{$\pm$610.0}} & -2900\tiny{$\pm$784.4} & -7379\tiny{$\pm$1510.} & -3036\tiny{$\pm$1024.} & -16977\tiny{$\pm$4269.} & -122\tiny{$\pm$746.8} & -59231\tiny{$\pm$8688.}\\
\cline{2-10}
&   &2 & \textbf{283\tiny{$\pm$26.21}} & 213\tiny{$\pm$32.72} & 247\tiny{$\pm$46.78} & 106\tiny{$\pm$18.96} & 255\tiny{$\pm$23.77} & \textbf{287\tiny{$\pm$25.67}} & -159\tiny{$\pm$128.3}\\
& REF-$q$ &7 & \textbf{3323\tiny{$\pm$194.7}} & 1886\tiny{$\pm$269.5} & 2949\tiny{$\pm$264.1} & 215\tiny{$\pm$221.7} & 2824\tiny{$\pm$214.2} & 2824\tiny{$\pm$297.9} & -26\tiny{$\pm$644.8}\\
&   &10 & \textbf{6340\tiny{$\pm$435.0}} & 2884\tiny{$\pm$575.9} & 5301\tiny{$\pm$612.5} & -230\tiny{$\pm$647.3} & 5229\tiny{$\pm$453.1} & 4981\tiny{$\pm$729.0} & 181\tiny{$\pm$1331.}\\
\cline{2-10}
&   &3 & 2567\tiny{$\pm$123.9} & 1708\tiny{$\pm$228.1} & \textbf{2742\tiny{$\pm$122.5}} & 1776\tiny{$\pm$124.7} & 2384\tiny{$\pm$105.7} & 2712\tiny{$\pm$107.7} & -3505\tiny{$\pm$719.7}\\
& TREA-$q$ &7 & 14249\tiny{$\pm$547.1} & 9627\tiny{$\pm$1154.} & 14473\tiny{$\pm$353.2} & 8342\tiny{$\pm$463.5} & 14364\tiny{$\pm$346.3} & \textbf{15436\tiny{$\pm$381.2}} & -17986\tiny{$\pm$2391.}\\
&   &10 & 29193\tiny{$\pm$1183.}  & 20437\tiny{$\pm$2679.} & 31878\tiny{$\pm$850.5} & 17306\tiny{$\pm$875.7} & 27100\tiny{$\pm$1061.} & \textbf{32421\tiny{$\pm$642.6}} & -32136\tiny{$\pm$5006.}\\
\midrule[1pt]
\multicolumn{3}{c|}{\makecell[c]{normalized\\mean performance}} & \textbf{9.802} & 9.104 & 6.486 & 8.091 & 4.762 & 9.080 & 1.990 \\
\bottomrule
\specialrule{0em}{1.5pt}{1.5pt}
\bottomrule
\end{tabular}}
\label{table_results_on_reward_settings}
\end{table*}

\noindent\textbf{Baselines.}~~
In order to verify the efficacy of distributional reward estimation in MARL with reward uncertainty,
we select the following methods as baselines: MADDPG~\cite{lowe2017multi} adopts a value-based centralized Q network to concatenate and process all agents' observations.
MAAC~\cite{iqbal2019actor} uses attention mechanism in CTDE paradigm.
MAPPO~\cite{yu2021surprising} reaches good results with on-policy method.
QMIX~\cite{rashid2018qmix} adopts monotonicity to decompose team reward.
IQL~\cite{tan1993multi} adopts completely independent training on each agent.
Apart from the above methods, we also compare with point-to-point (p2p-MARL) reward estimation~\cite{romoff2018reward} and global reward estimation (GRE-MARL).

\noindent\textbf{Metrics.}\label{metrics}~~~~For all methods we consider, we report the final performance under different reward settings ($r_{dete}$,$r_{dist}$,$r_{ac-dist}$), which are introduced above.
Following prior studies~\cite{iqbal2019actor,lowe2017multi}, we set the episode length as 25 for all experiments and use the mean episode reward as the evaluation metric. 
Because the value of performance in different scenarios varies considerably, we also consider performance normalization based on all methods.
For example, if we have performance values $(M_1, M_2, M_3)$ of three models, then the normalized performance is defined by $M_v^{\prime}=\frac{\omega(M_v-min(M_1, M_2, M_3))}{max(M_1, M_2, M_3)-min(M_1, M_2, M_3)}$, where $\omega$ enables us to distinguish the differences of methods appropriately. We set $\omega=10$ in the experiments. 
More details about training and implementation are provided in Appendix~\ref{Implement Details}.

\begin{figure*}[t!]
\vspace{-1em}
 \begin{center}
 \includegraphics[angle=0,width=0.99\textwidth]{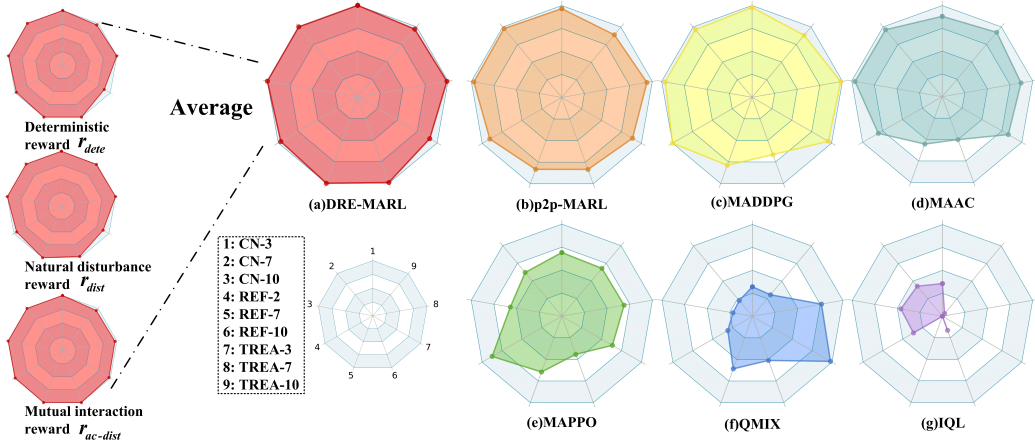}
 \caption{The normalized mean performance under different reward settings ($r_{dete}$, $r_{dist}$, $r_{ac-dist}$). Each polygon (shown right) is averaged on three sub-polygons (shown left), where each sub-polygon represents the corresponding reward setting. Each vertex of the sub-polygon denotes the normalized performance, which matches clockwise from CN-3 to TREA-10. The detailed calculation method of normalized performance can be found in Section \ref{metrics}.}
 \label{rader_fig}
 \end{center}
 \vspace{-1em}
 \end{figure*}

\subsection{Results and Analysis}\label{Results and Analysis} 
Table~\ref{table_results_on_reward_settings} shows the performance of different methods with the three reward settings.
We report the true value of each experiment.
The results show that in most of the environments, DRE-MARL reaches the best performance, which reveals the effectiveness of distributional reward estimation.
In several scenarios, such as TREA-$7$ with $r_{dete}$ and $r_{dist}$, DRE-MARL is not the best, but its performance is very close to the best.
Under the $r_{ac-dist}$ reward uncertainty, DRE-MARL fails in TREA-$3$, TREA-$7$, and TREA-$10$.
That is because $r_{ac-dist}$ brings more difficulties for attention-based methods in TREA: DRE-MARL and MAAC all adopt multi-head attention architecture, and they are subject to the same trend of changing.
We find that in $r_{dete}$ and $r_{dist}$ settings, MAAC achieves the competitive performance, while our method reaches the best performance.
But in the $r_{ac-dist}$ setting, the performance of MAAC is pretty bad, and also, DRE-MARL is not the best one.
If we consider the utility of DRE based on attention in the setting of TREA with $r_{ac-dist}$ (i.e., Comparation between DRE-MARL and MAAC), DRE brings at least $50\%$ improvement. 

To reveal the comprehensive abilities of the above methods, we normalize the performances of all methods and plot the overall performances as shown in Figure~\ref{rader_fig}.
The vertexes (1$\sim$9) of the polygon represent the corresponding scenarios (CN-$q$,REF-$q$,TREA-$q$).
To get the values on vertexes of polygon, we average the normalized values of performances under all reward settings $r_{dete}$,$r_{dist}$,$r_{ac-dist}$.
Figure~\ref{rader_fig} suggests that DRE-MARL consistently achieves the best overall performance in almost all scenarios.
In REF-$2$ and TREA-$3$, the performance of DRE-MARL is just slightly lower than MADDPG because the agent number is small, and it is relatively easy to train.
Additionally, The results also illustrate that DRE-MARL has a more robust performance with respect to various reward uncertainties because the DRE can capture the reward uncertainty and stabilize training under different reward settings.
Besides, exhaustive experiments are shown in Appendix~\ref{Additional Experiments}.

\subsection{Ablation Study}

\begin{figure*}[t!]
\hspace{-1em}
 \begin{center}
 \includegraphics[angle=0,width=1\textwidth]{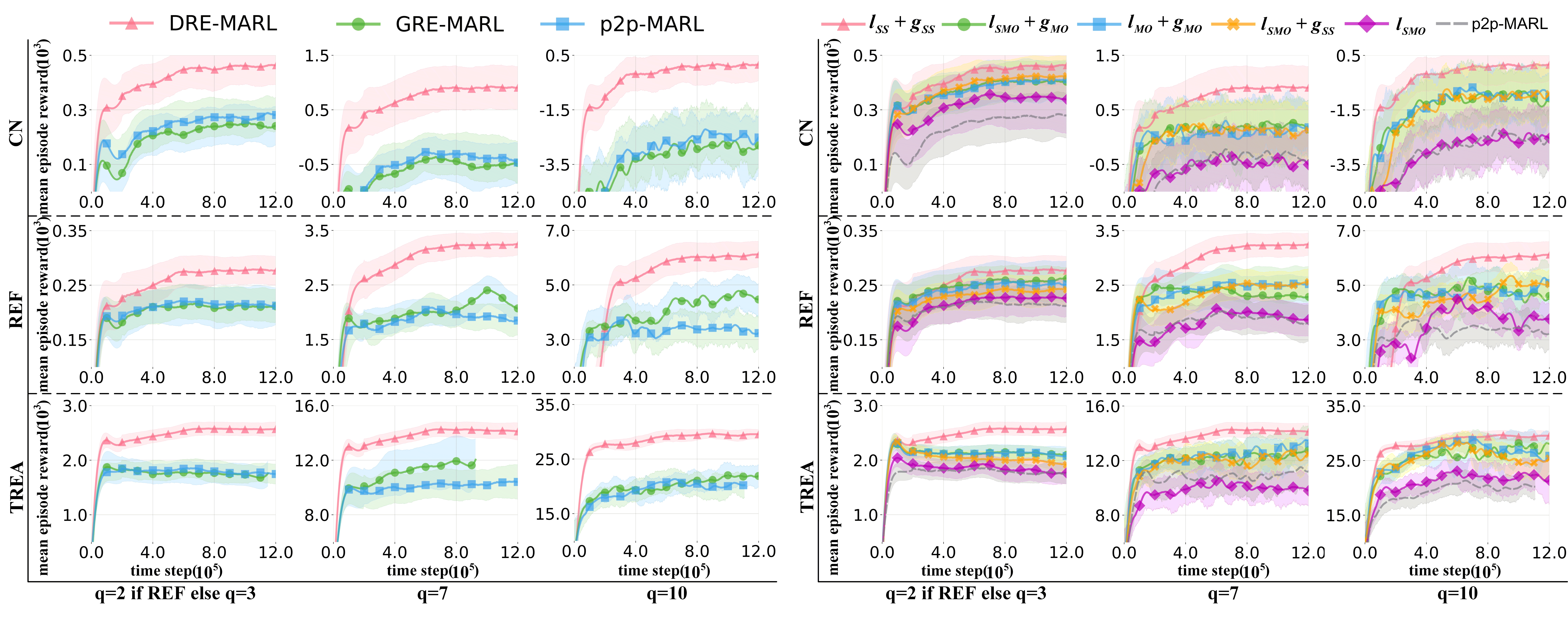}
 \caption{\emph{Left}: Ablation study under reward setting $r_{ac-dist}$. We change the reward estimation strategy, such as global reward estimation (GRE-MARL) and point-to-point reward estimation (p2p-MARL), and probe the influence of different reward estimation methods. \emph{Right}: Reward aggregation analysis on different aggregated schemes. We evaluate our model with different and frequently-used reward aggregation strategies, which enable us to assess our method comprehensively.}
 \label{abalation}
 \end{center}
 \vspace{-1em}
 \end{figure*}

Firstly, we perform the ablation study to investigate the impact of different reward estimation methods on the proposed model:
1)DRE-MARL: distributional reward estimation. 2) GRE-MARL: global reward estimation based on joint state-action pairs. 3) p2p-MARL: point-to-point reward estimation.
As shown in Figure~\ref{abalation} (left), we report the learning curves of the above methods under the reward setting of $r_{ac-dist}$.
The results show that DRE-MARL is able to achieve better asymptotic performance under different scenarios.
GRE-MARL fails to capture the reward uncertainty because the joint state-action space of MARL is huge, which brings more difficulties for global reward estimation.
p2p-MARL does not consider the reward uncertainty caused by mutual interactions among agents as the one-to-many mapping problem, so it performs badly by adopting the one-to-one reward estimation method in MARL scenarios. 

To probe the effect of different reward aggregation schemes on the performance, we report the performance of several aggregation schemes in Figure~\ref{abalation} (right) under the $r_{ac-dist}$ setting.
Recall the reward aggregation introduced in Section~\ref{aggregation}, we select the following aggregation schemes. 1) $l_{SS}+g_{SS}$, 2) $l_{SMO}+g_{MO}$, 3) $l_{MO}+g_{MO}$, 4) $l_{SMO}+g_{SS}$, 5) $l_{SMO}$ (no aggregation).
In Figure~\ref{abalation} (right), we remark that the utility of reward aggregation, which makes the agent make decisions thoughtfully, consistently improves performance in various scenarios compared with GRE-MARL and p2p-MARL. 
Among all aggregation schemes, $l_{SS}+g_{SS}$ reaches the best performance, and the reported results are also based on this aggregation scheme. More details are shown in Appendix~\ref{Additional aggregation analysis}.

\vspace{-0.1em}
\section{Conclusion, Limitations, and Broader Impact}\label{conclusion}
In this paper, we present DRE-MARL, a general and effective reward estimation method for reward uncertainty in MARL. 
To capture the reward uncertainty and stabilize the training process, we investigate the benefits of distributional reward estimation followed by reward aggregation in MARL when reward uncertainty is present. 
The intuition is that careful consideration of all possible consequences is useful for learning policies. 
For our proposed DRE-MARL, we propose two stages to perform distributional reward estimation:  
we first design the multi-action-branch reward estimation to model reward distributions on all action branches. 
Then we propose policy-weighted reward aggregation with environment rewards and sampled rewards. 
Here we emphasize that reward aggregation augments the reward from one action branch to all action branches and thus provides more precise updating signals for the critic and actors.
Experimental results verify that agents learned by our framework can achieve the best performance and show strong robustness on different types of reward uncertainty. 
In conclusion, we hope that our work could demonstrate the potential of distributional reward estimation to improve the capacity of MARL when reward uncertainty is present and encourage future works to develop novel reward estimation methods.

In terms of limitations, DRE-MARL requires a prior assumption about the form of reward distribution. 
Just like tanglesome signals can be factorized into the superposition of simple but elementary sinusoidal signals, we may consider using a cluster of basic distributions as reward distribution in future work.
Another limitation is that reward aggregation is only used in discrete action space. Two solutions may be investigated in the future to solve this limitation: 1) Discretizing of the range of the available action value. 2) Learning a network with actions as inputs, just like how we convert discrete-action DQN~\cite{mnih2013playing} into continuous-action Q network in DDPG~\cite{lillicrap2015continuous}.
We do not see any negative societal impacts of our work while using our method in practice.

\section*{Acknowledgments and Disclosure of Funding}
We thank all anonymous reviewers for their constructive suggestions.
This work is partially supported in part by the National Natural Science Foundation of China under grants Nos.61902145, 61976102, and U19A2065; the International Cooperation Project under grant No. 20220402009GH; and the National Key R$\&$D Program of China under grants Nos. 2021ZD0112501 and 2021ZD0112502. 

\bibliography{reference}
\bibliographystyle{plainnat}

\appendix


\clearpage

\Large
\textbf{Appendix}

\normalsize
\section{Pseudocode of DRE-MARL}\label{algorithm}
The pseudocode for DRE-MARL training is shown in Algorithm~\ref{DRE-MARL algorithm}, which takes the following steps. 1) We perform several interactions with the environment and collect experiences in advance. 2) As shown in lines $4-9$ (Algorithm~\ref{DRE-MARL algorithm}), we collect transitions and deposit them in replay buffer $D$. 3) During the training process, i.e., lines $10-17$ (Algorithm~\ref{DRE-MARL algorithm}), we update the centralized critic, decentralized actors, and reward estimators every 100 time steps following previous methods~\cite{iqbal2019actor, lowe2017multi} when the episode ends.
4) We will evaluate our model and refresh the replay buffer periodically.
Source code is available at https://github.com/JF-Hu/DRE-MARL.git.

\begin{algorithm}[h!]
  \LinesNumbered
  \caption{Multi-agent Distributional Reward Estimation (DRE-MARL)}
  \KwIn{$N$ policies $\{\pi^{i}\}_{i\in\mathcal{N}}$ and target policies $\{\tilde{\pi}^{i}\}_{i\in\mathcal{N}}$ parameterized by $\boldsymbol{\theta}=\{\theta_i\}_{i\in\mathcal{N}}$ and $\boldsymbol{\tilde{\theta}}=\{\tilde{\theta}_i\}_{i\in\mathcal{N}}$, respectively; Centralized critic and target centralized critic parameterized by $\psi$ and $\tilde{\psi}$, respectively; $N$ reward estimators parameterized by $\bm{\varphi}=\{\varphi_{i}\}_{i\in\mathcal{N}}$; The environment with reward uncertainty}
  \KwOut{$\psi$, $\boldsymbol{\theta}$, $\bm{\varphi}$}
  \textbf{Initialize:}$\psi$, $\tilde{\psi}$, $\boldsymbol{\theta}$, $\boldsymbol{\tilde{\theta}}$, $\bm{\varphi}$, reply buffer $D$\\
  Pre-interact with environment\\
  \For{each iteration}{
    \For{each time step $t$}{
        Sample action $a^{i}_{k,t}\sim\pi^{i}(\cdot|o^{i}_{t})$ for each agent $i$\\
        Execute joint action $\boldsymbol{a_t}$ in the environment\\
        Observe next observation $o^{i}_{t+1}$ and reward $r^{i}_{k,t}$ of each agent $i$\\
        Store $\left(\{o^{i}_{t}\},\{\pi^{i}(\cdot|o^{i}_{t})\},\{a^{i}_{k,t}\},\{r^{i}_{k}\},\{o^{i}_{t+1}\}\right)$ in replay buffer $D$\\
    }
    \If{it's time to update}{
        Sample a batch of transitions $\mathcal{B}$ from $D$\\
        Update distributional reward estimators with Equation~\ref{reward update}\\
        Sampling rewards $\bm{\hat{r}^{i}}$ from multi-action-branch reward distributions $\widehat{R}^{i}$\\
        Perform reward aggregation with Equation~\ref{reward aggregation R} and Equation~\ref{reward aggregation r}\\
        Update centralized critic and decentralized actors\\
        Update target networks $\{{\pi}_{\Tilde{\theta}_{i}}\}_{i\in\mathcal{N}}$, $V_{\gamma,\Tilde{\psi}}$: $\tilde{\theta_{i}}\leftarrow\tau\theta_{i}+(1-\tau)\tilde{\theta_{i}}$,$\tilde{\psi}\leftarrow\tau\psi+(1-\tau)\tilde{\psi}$\\
    }
    Evaluate individual policies periodically\\
    Refresh replay buffer periodically\\
  }
\label{DRE-MARL algorithm}
\end{algorithm}

\section{Additional Experiments}\label{Additional Experiments}
\subsection{Environmental description}\label{Environmental description}
\textbf{Cooperation Navigation.}~~In this environment, three agents must collaborate to cover all landmarks and avoid colliding with each other simultaneously.
The property of the received reward in this environment is set to be collaborative.
For more complex environmental settings, we increase the number of agents and landmarks (i.e., Evaluation on  7 and 10 agents).
We use the abbreviation CN to denote this environment.

\textbf{Reference.}~~It is a scenario with two agents and three landmarks. The difference between Cooperative Navigation and Reference is that the target landmark of each agent is only known to its partner.
Thus every agent must convey correct information to each other to accomplish the task.
To evaluate the capacity of our method in complicated environments, likewise the Cooperative Navigation, we also select 7 and 10 agents for evaluation.
We use the abbreviation REF to denote this environment.

\begin{figure*}[t!]
\hspace{-1em}
 \begin{center}
 \includegraphics[angle=0,width=0.8\textwidth]{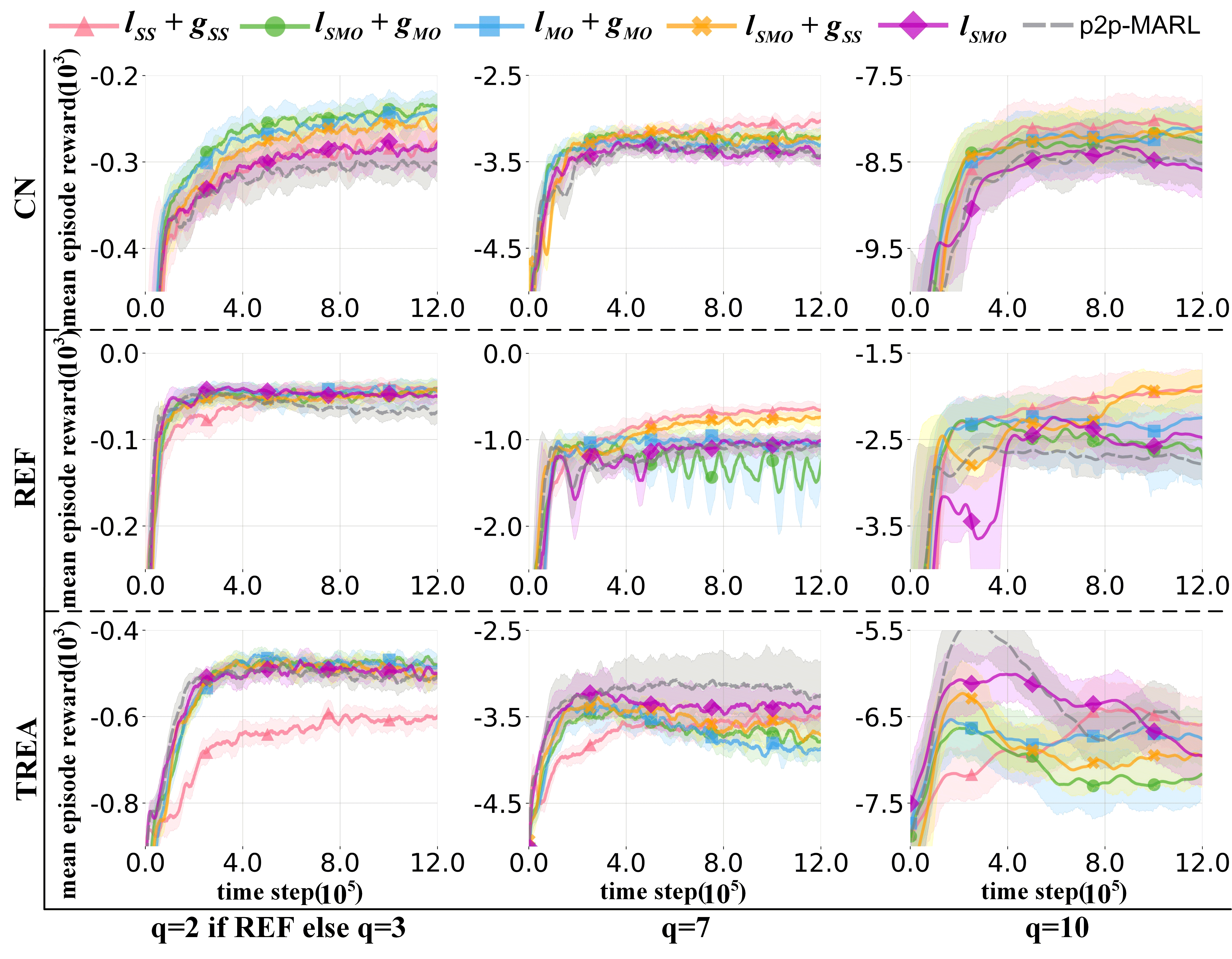}
 \caption{Comparison of the performance on different aggregated schemes while training with the $r_{ac-dist}$ setting and evaluating without the $r_{ac-dist}$ setting.}
 \label{Appendix_figure4}
 \end{center}
 \vspace{-1em}
 \end{figure*}

\begin{figure*}[t!]
\hspace{-1em}
 \begin{center}
 \includegraphics[angle=0,width=0.8\textwidth]{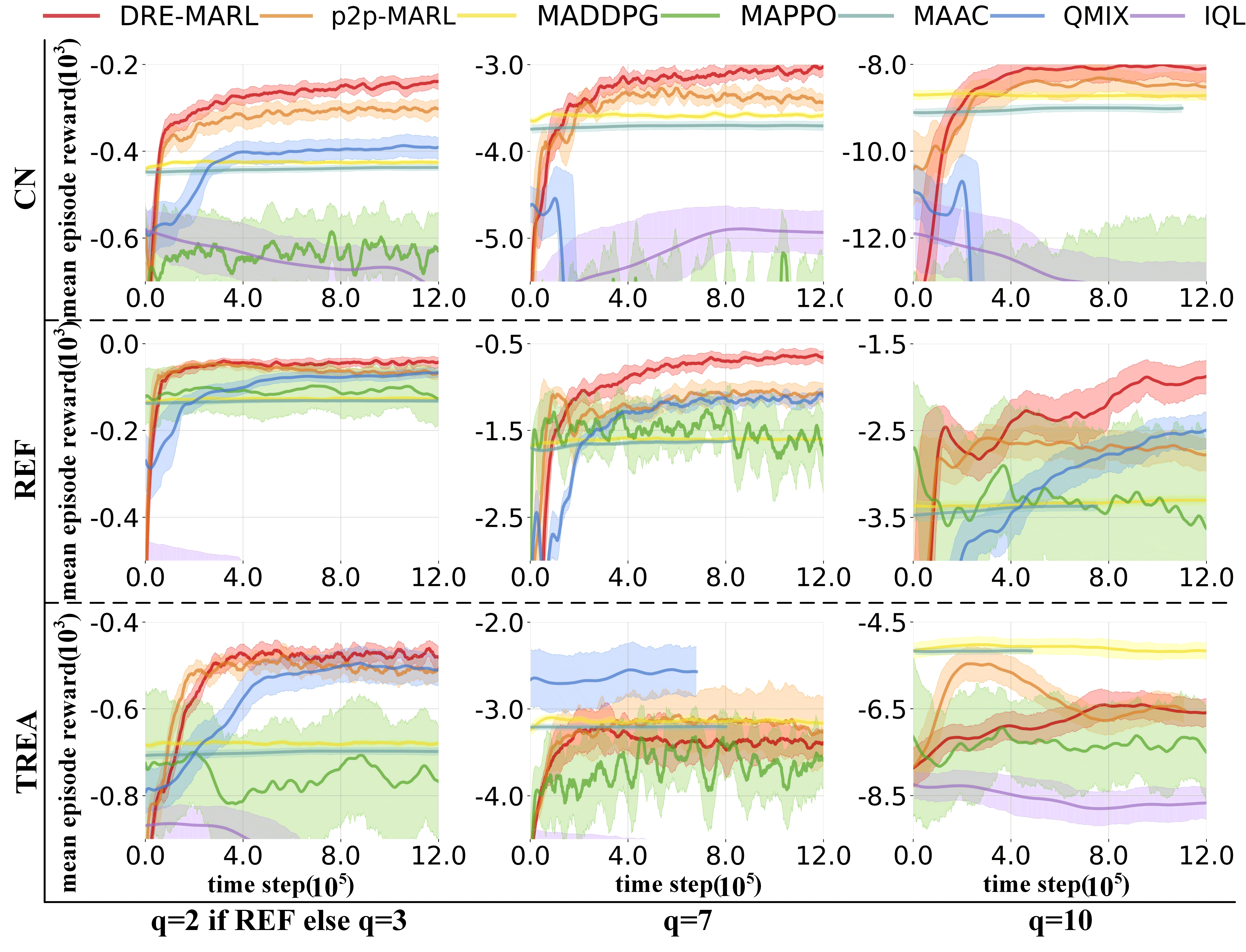}
 \caption{Comparison of DRE-MARL and different baselines while training with the $r_{ac-dist}$ setting and evaluating without the $r_{ac-dist}$ setting.}
 \label{Appendix_figure5}
 \end{center}
 \vspace{-1em}
 \end{figure*}

\textbf{Treasure Collection.}~~In this scenario, a successful process of treasure collection contains two stages: the first stage is that collectors cover the correct landmarks (i.e., collecting treasure), and the second is that the bank agents successfully receive the treasure.
In this scenario, the per episode length is set to 25 because fewer steps bring more challenges to accomplish the two-stage task. The number of collectors and banks is the same, so the collectors should find their unique bank from all banks.
Analogously, we choose 3c3b, 7c7b, 10c10b as measurements.

\begin{table*}[t]
\centering
\small
\renewcommand{\arraystretch}{1.5}
\caption{Performance comparison of DRE-MARL, DRE-MARL variants, and several SOTA MARL algorithms while training with the individual rewards under the $r_{ac-dist}$ setting. The values represent mean episodic rewards.}
\resizebox{\textwidth}{!}{
\begin{tabular}{l|r|r|r|r|r|r|r}
\toprule
\specialrule{0em}{1.5pt}{1.5pt}
\toprule
 Reward Setting & \multicolumn{6}{c|}{\large{$r_{ac-dist}$}} & \multirow{4}{*}{\makecell[c]{Normalized\\Performance}}\\
 \cmidrule{1-7}
 Scenario & \multicolumn{3}{c|}{CN-$q$} & \multicolumn{3}{c|}{REF-$q$} & \\
 \cmidrule{1-7}
 $q$ & 3& 7& \multicolumn{1}{c|}{10} & 2& 7& 10 & \\
 \cmidrule{1-8}
 DRE-MARL $(l_{SS}+g_{SS})$   & 129\tiny{$\pm$26.93} & 36\tiny{$\pm$77.89}  & -122\tiny{$\pm$113.8} & 23\tiny{$\pm$64.11}  & -13\tiny{$\pm$201.3}  & 402\tiny{$\pm$108.3} & 7.893\\
 DRE-MARL $(l_{SMO}+g_{MO})$  & 126\tiny{$\pm$26.54} & 11\tiny{$\pm$75.25}  & -172\tiny{$\pm$116.4} & 82\tiny{$\pm$40.48}  & 258\tiny{$\pm$68.25}  & 278\tiny{$\pm$83.82} & 8.467\\
 DRE-MARL $(l_{MO}+g_{MO})$   & 144\tiny{$\pm$22.80} & 43\tiny{$\pm$66.72}  & -163\tiny{$\pm$78.67} & 113\tiny{$\pm$17.16} & 310\tiny{$\pm$55.84}  & 454\tiny{$\pm$72.59} & 9.360\\
 DRE-MARL $(l_{SMO}+g_{SS})$  & 133\tiny{$\pm$22.79} & 39\tiny{$\pm$70.88}  & -86\tiny{$\pm$113.8}  & 11\tiny{$\pm$61.19}  & 177\tiny{$\pm$120.8}  & 74\tiny{$\pm$133.6}  & 7.904\\
 DRE-MARL $(l_{SMO})$         & 99\tiny{$\pm$24.99}  & 1\tiny{$\pm$69.99}   & -186\tiny{$\pm$120.6} & 44\tiny{$\pm$45.23}  & 171\tiny{$\pm$87.49}  & 300\tiny{$\pm$89.35} & 7.737\\
 p2p-MARL                     & 92\tiny{$\pm$25.27}  & -15\tiny{$\pm$72.02} & -171\tiny{$\pm$116.9} & 35\tiny{$\pm$50.32}  & 192\tiny{$\pm$101.8}  & 227\tiny{$\pm$87.09} & 7.517\\
 GRE-MARL                     & 113\tiny{$\pm$26.66} & -18\tiny{$\pm$60.79} & -249\tiny{$\pm$98.28} & 116\tiny{$\pm$16.30} & 244\tiny{$\pm$67.66}  & 231\tiny{$\pm$98.08} & 8.171\\
 MADDPG                       & 138\tiny{$\pm$30.59} & 47\tiny{$\pm$84.41}  & -217\tiny{$\pm$125.42}& -108\tiny{$\pm$102.8}& -371\tiny{$\pm$166.0} & -609\tiny{$\pm$196.7}& 4.608\\
 MAPPO                        & 10\tiny{$\pm$43.89}  & -313\tiny{$\pm$168.2}& -686\tiny{$\pm$299.0} & -50\tiny{$\pm$89.86} & -103\tiny{$\pm$243.2} & -61\tiny{$\pm$312.4}  & 2.457\\
 MAAC                         & 132\tiny{$\pm$17.02} & 34\tiny{$\pm$47.49}  & -146\tiny{$\pm$70.89} & 11\tiny{$\pm$25.14}  & 26\tiny{$\pm$53.52}   & 31\tiny{$\pm$66.36}   & 7.324\\
 QMIX                         & -4\tiny{$\pm$70.57}  & -209\tiny{$\pm$170.6}& -715\tiny{$\pm$195.83}& 125\tiny{$\pm$15.78} & 380\tiny{$\pm$42.38}  & 577\tiny{$\pm$45.12}  & 5.970\\
 IQL                          & -66\tiny{$\pm$42.65} & -309\tiny{$\pm$115.3}& -705\tiny{$\pm$151.5} & -62\tiny{$\pm$76.95} & -210\tiny{$\pm$137.1} & -271\tiny{$\pm$173.8} & 1.204\\
\bottomrule
\specialrule{0em}{1.5pt}{1.5pt}
\bottomrule
\end{tabular}
}
\label{Appendix_table2}
\end{table*}

\begin{figure*}[t]
\hspace{-1em}
 \begin{center}
 \includegraphics[angle=0,width=1\textwidth]{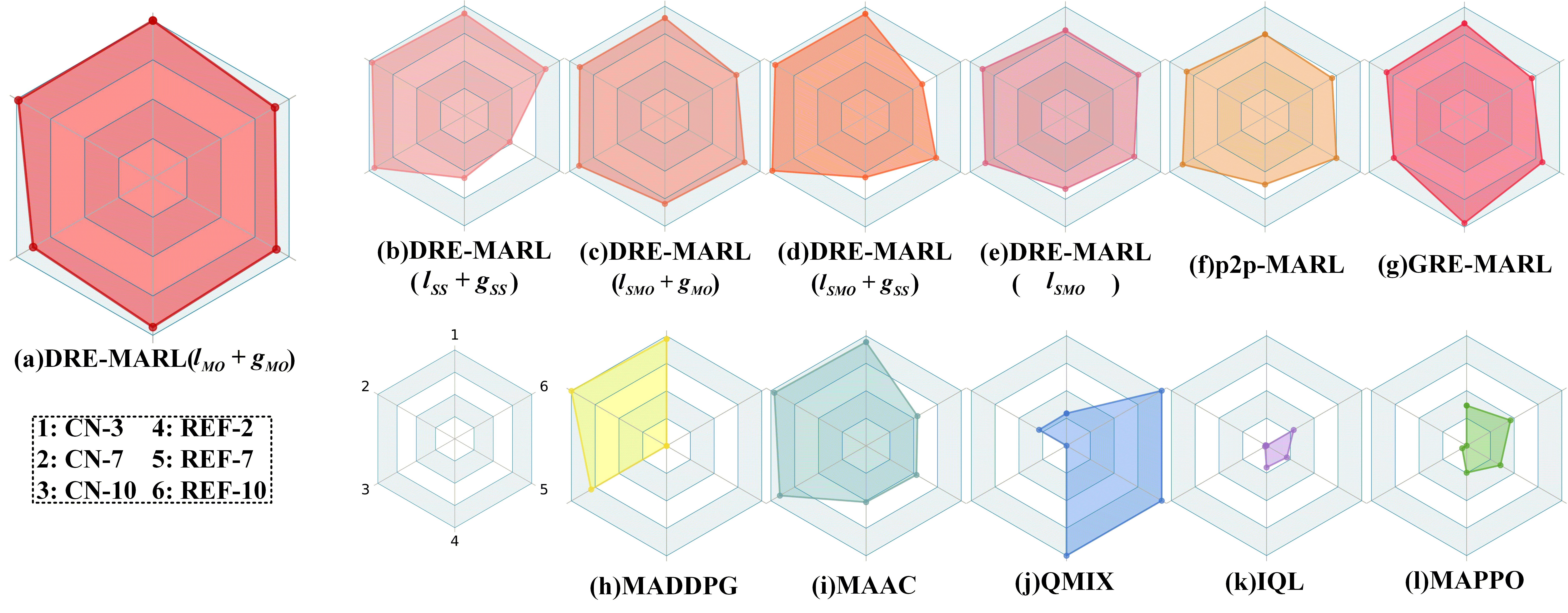}
 \caption{The normalized performance while training with the individual rewards and evaluating with the $r_{ac-dist}$ setting.}
 \label{Appendix_figure6}
 \end{center}
 \end{figure*}

\subsection{Additional aggregation analysis}\label{Additional aggregation analysis}
There are two choices when evaluating the proposed method. 1) Evaluating in the environment with the reward uncertainty setting as same as training. 2) Evaluating the environment without the setting of reward uncertainty.
We usually select the first choice in practice because the training and testing environments are the same. The second choice is also important because it can show whether our model is stuck in the setting of reward uncertainty.
We have report the results of first choice in Table~\ref{table_results_on_reward_settings}, Figure~\ref{rader_fig}, and Figure~\ref{abalation}.
In the next parts, we report the performance of different reward aggregation schemes and other baselines based on the second choice.

We evaluate the performance under the team reward setting in the environments where the $r_{ac-dist}$ is not added.
As shown in Figure~\ref{Appendix_figure4}, in most environments, the proposed reward aggregation schemes achieve better performance than p2p-MARL.
The aggregation scheme $l_{SMO}+g_{SS}$ is better than $l_{SS}+g_{SS}$ while Figure~\ref{abalation} (right) shows that $l_{SS}+g_{SS}$ is the best.
This enlightens us different aggregation schemes are suitable for various purposes.
In other words, if we want to train a model in precise environments designed by humans, we should choose $l_{SMO}+g_{SS}$ rather than $l_{SS}+g_{SS}$.
But if we are given an environment with reward uncertainty and the evaluation is also performed in it, the aggregation scheme $l_{SS}+g_{SS}$ is a better choice.

Furthermore, we also evaluate the performance while training with individual rewards, i.e., the agents can only receive the individual rewards rather than the team rewards during the training process.
This environment setting is more complicated than the team reward setting because the agents can cause invalid updating on the centralized critic.
The reward aggregation can provide a consistent updating direction for the centralized critic, so as illustrated in Table~\ref{Appendix_table2} and Table~\ref{Appendix_table3}, our method achieves a better performance than the other baselines.
The aggregation scheme $l_{MO}+g_{MO}$ reaches the best performance because the mean operation offers a relatively consistent direction of updating.

\begin{table*}[t]
\centering
\small
\renewcommand{\arraystretch}{1.5}
\caption{Performance comparison of DRE-MARL, DRE-MARL variants, and several SOTA MARL algorithms while training with the individual rewards and evaluating without the $r_{ac-dist}$ setting. The values represent mean episodic rewards.}
\resizebox{\textwidth}{!}{
\begin{tabular}{l|r|r|r|r|r|r|r}
\toprule
\specialrule{0em}{1.5pt}{1.5pt}
\toprule
 Reward Setting & \multicolumn{6}{c|}{\large{$r_{ac-dist}$}} & \multirow{4}{*}{\makecell[c]{Normalized\\Performance}}\\
 \cmidrule{1-7}
 Scenario & \multicolumn{3}{c|}{CN-$q$} & \multicolumn{3}{c|}{REF-$q$} & \\
 \cmidrule{1-7}
 $q$ & 3& 7& \multicolumn{1}{c|}{10} & 2& 7& 10 & \\
 \cmidrule{1-8}
 DRE-MARL $(l_{SS}+g_{SS})$   & -403\tiny{$\pm$33.82} & -4148\tiny{$\pm$416.8}  & -9283\tiny{$\pm$442.9} & -358\tiny{$\pm$112.2}  & -5663\tiny{$\pm$2102.}  & -4711\tiny{$\pm$996.3}  & 7.016\\
 DRE-MARL $(l_{SMO}+g_{MO})$  & -320\tiny{$\pm$32.84} & -3904\tiny{$\pm$204.1}  & -9775\tiny{$\pm$396.8} & -166\tiny{$\pm$81.60}  & -2183\tiny{$\pm$300.2}  & -5744\tiny{$\pm$668.7}  & 8.457\\
 DRE-MARL $(l_{MO}+g_{MO})$   & -230\tiny{$\pm$15.54} & -3260\tiny{$\pm$126.8}  & -8491\tiny{$\pm$223.9} & -70\tiny{$\pm$13.92}   & -1361\tiny{$\pm$175.7}  & -2882\tiny{$\pm$384.3}  & 9.797\\
 DRE-MARL $(l_{SMO}+g_{SS})$  & -283\tiny{$\pm$23.55} & -3587\tiny{$\pm$158.7}  & -8791\tiny{$\pm$388.5} & -310\tiny{$\pm$113.5}  & -2825\tiny{$\pm$800.7}  & -7847\tiny{$\pm$1756.}  & 7.920\\
 DRE-MARL $(l_{SMO})$         & -350\tiny{$\pm$22.56} & -3672\tiny{$\pm$161.9}  & -9543\tiny{$\pm$428.2} & -195\tiny{$\pm$49.99}  & -2161\tiny{$\pm$382.7}  & -4421\tiny{$\pm$536.6}  & 8.574\\
 p2p-MARL                     & -367\tiny{$\pm$23.72} & -3717\tiny{$\pm$160.7}  & -9241\tiny{$\pm$397.3} & -212\tiny{$\pm$51.14}  & -2213\tiny{$\pm$452.2}  & -4315\tiny{$\pm$489.4}  & 8.591\\
 GRE-MARL                     & -467\tiny{$\pm$29.11} & -4522\tiny{$\pm$165.5}  & -11408\tiny{$\pm$351.5}& -103\tiny{$\pm$16.75}  & -2604\tiny{$\pm$435.7}  & -6964\tiny{$\pm$773.8}  & 7.715\\
 MADDPG                       & -429\tiny{$\pm$6.242} & -3589\tiny{$\pm$42.26}  & -8693\tiny{$\pm$66.01} & -133\tiny{$\pm$3.568}  & -1635\tiny{$\pm$21.71}  & -3348\tiny{$\pm$35.76}  & 8.986\\
 MAPPO                        & -1074\tiny{$\pm$240.4}& -9349\tiny{$\pm$2315.}  & -21475\tiny{$\pm$5276.}& -603\tiny{$\pm$216.0}  & -6357\tiny{$\pm$2021.}  & -11532\tiny{$\pm$3721.} & 1.007\\
 MAAC                         & -441\tiny{$\pm$5.511} & -3732\tiny{$\pm$35.19}  & -9076\tiny{$\pm$77.81} & -140\tiny{$\pm$2.933}  & -1712\tiny{$\pm$23.21}  & -3504\tiny{$\pm$40.08}  & 8.817\\
 QMIX                         & -671\tiny{$\pm$234.8} & -4870\tiny{$\pm$1128.}  & -14964\tiny{$\pm$3096.}& -66\tiny{$\pm$10.64}   & -1128\tiny{$\pm$100.0}  & -2327\tiny{$\pm$145.4}  & 7.790\\
 IQL                          & -973\tiny{$\pm$38.33} & -6053\tiny{$\pm$918.8}  & -13836\tiny{$\pm$1331.}& -631\tiny{$\pm$75.07}  & -7905\tiny{$\pm$381.7}  & -16048\tiny{$\pm$744.0} & 2.081\\
\bottomrule
\specialrule{0em}{1.5pt}{1.5pt}
\bottomrule
\end{tabular}
}
\label{Appendix_table3}
\end{table*}

\begin{figure*}[t]
\hspace{-1em}
 \begin{center}
 \includegraphics[angle=0,width=1\textwidth]{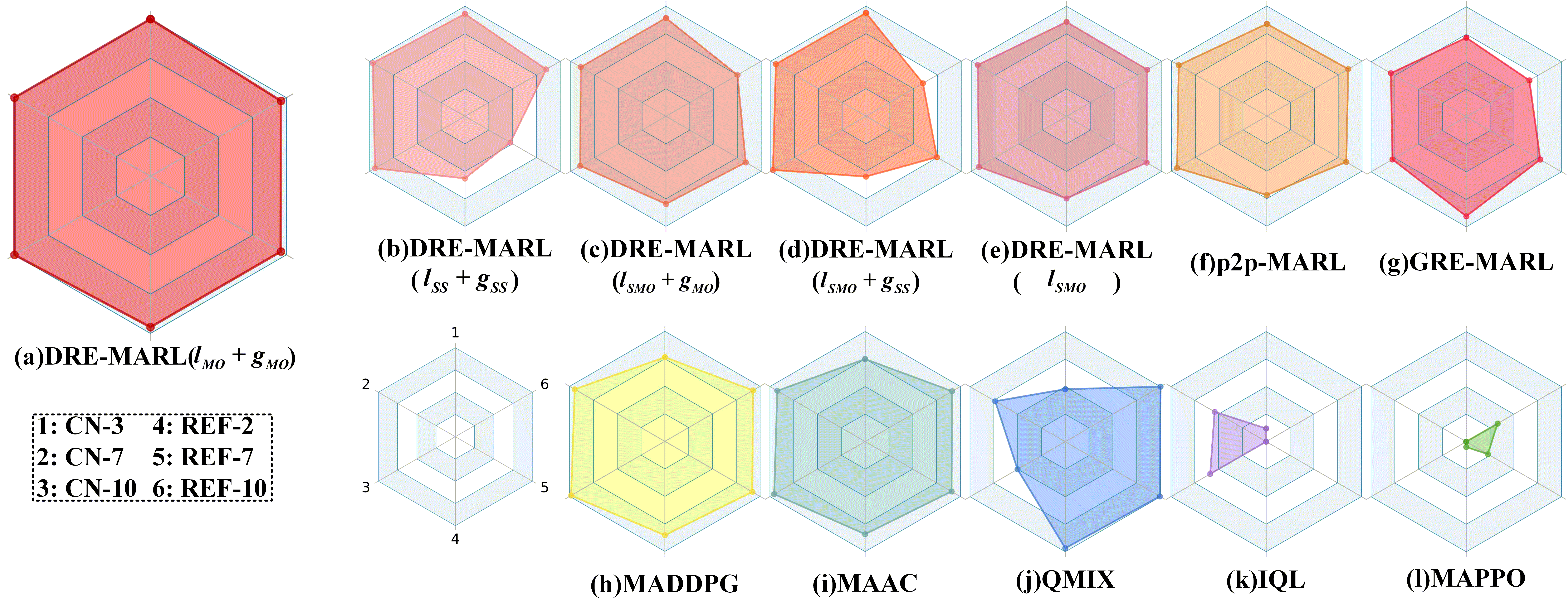}
 \caption{The normalized performance while training with the individual rewards and evaluating without the $r_{ac-dist}$ setting.}
 \label{Appendix_figure7}
 \end{center}
 \vspace{-1em}
 \end{figure*}

\subsection{Additional experiment results}\label{Additional experiment results}

In order to verify whether our model will be stuck in the setting of reward uncertainty, we evaluate the proposed method and various baselines in the environments without reward uncertainty.
We report the learning curves in Figure~\ref{Appendix_figure5}.
DRE-MARL performs better than different baselines, which illustrates that DRE-MARL can reduce the impact of reward uncertainty by adopting multi-action-branch reward estimation and policy-weighted reward aggregation during the training process.

Multi-action-branch reward estimation followed by reward aggregation not only provides a solution to reduce reward uncertainty by augmenting reward from one action branch to all action branches but also offers a good way to improve collaboration by considering the aggregated rewards of all agents.
In order to evaluate the performance just given the individual rewards, we set the environment to be non-collaborative during the training process, i.e., the agents can only receive their individual rewards rather than the team reward.
Table~\ref{Appendix_table2} and Table~\ref{Appendix_table3} show the performance of baselines and different aggregation schemes in several scenario variants.
The comprehensive abilities of the above methods are shown in Figure~\ref{Appendix_figure6} and Figure~\ref{Appendix_figure7}.
DRE-MARL ($l_{MO}+g_{MO}$) achieves better performance than other aggregation schemes and baselines in most environments.
The reason is that reward aggregation provides a good solution to consider the holistic impact of individual agents when updating the centralized critic so as to achieve the common goal.
Although QMIX reaches higher performance in REF, the difference is slight, and the overall performance of DRE-MARL ($l_{MO}+g_{MO}$) is obviously better than QMIX.
Besides, in order to train the QMIX model, we actually calculate the team reward for QMIX during training, which violates the environment setting of training with the individual rewards.
The value decomposition adopted by QMIX can be regarded as one backward method from the team reward to the individual rewards for solving MARL tasks.
Our method provides a brand-new forward solution from the individual rewards to the team reward by using multi-action-branch reward estimation followed by policy-weighted reward aggregation.

\section{Implement Details}\label{Implement Details}

\begin{figure*}[t!]
\hspace{-1em}
 \begin{center}
 \includegraphics[angle=0,width=0.8\textwidth]{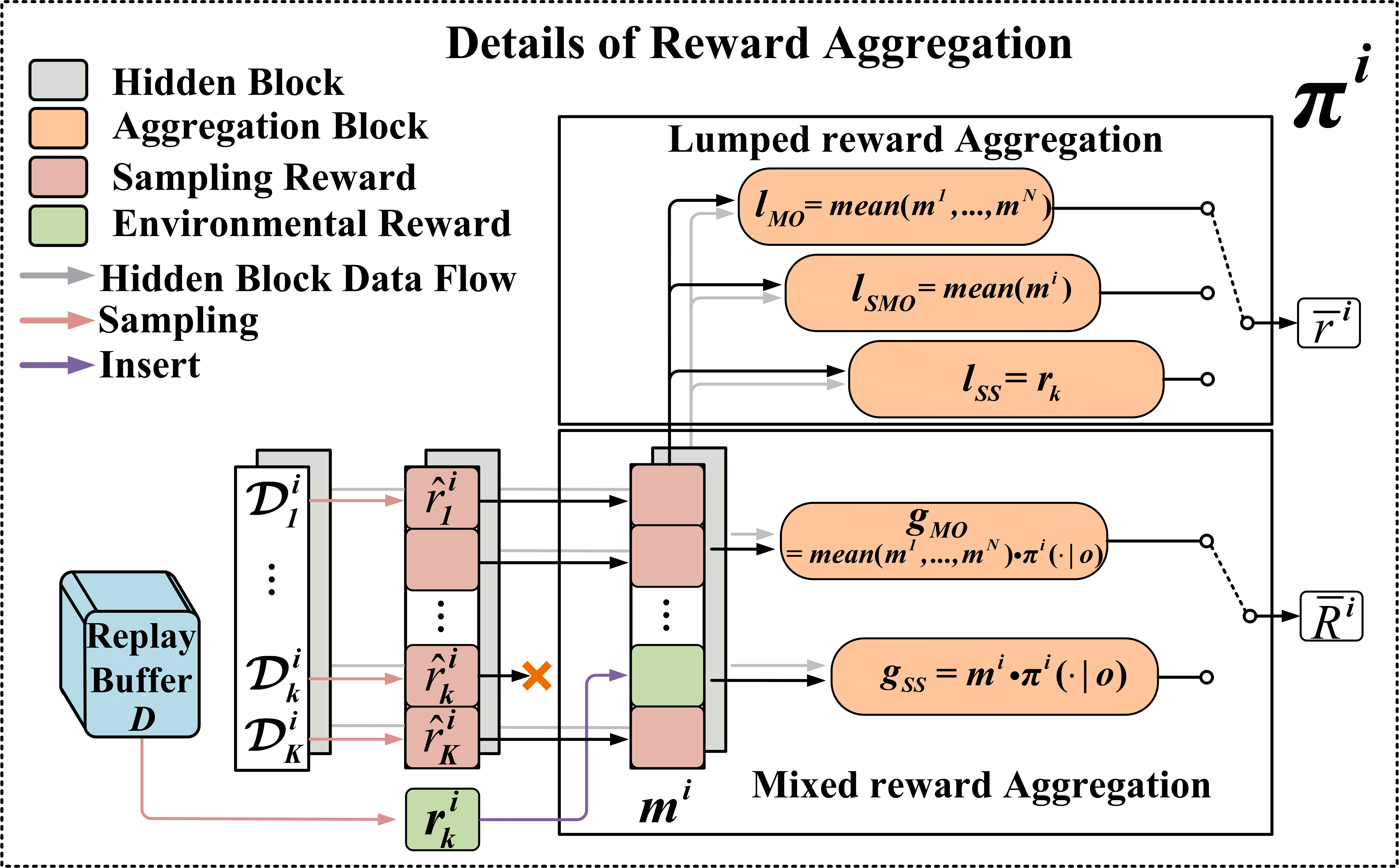}
 \caption{Graphical description of reward aggregation.}
 \label{reward aggregation}
 \end{center}
 \vspace{-1em}
 \end{figure*}

\noindent{\textbf{Details of Reward Aggregation.}}~~~~
For each agent, we first sample rewards $\hat{r}^{i}$ from estimated reward distributions $\mathcal{D}^{i}$.
Then, we construct the built-up reward vector $m^{i}$ by replacing the value of $\hat{r}^{i}$ on the $k$-th action branch with the environment reward $r^{i}_{k}$.
Next, we aggregate $\{m^{i}\}_{i\in\mathcal{N}}$ with different operations, as shown in Figure~\ref{reward aggregation}.
Finally, we obtain the mixed reward $\Bar{R}$ and the lumped reward $\Bar{r}$ to train the centralized critic and the decentralized actors, respectively.
The details of reward aggregation are shown in Figure~\ref{reward aggregation}.

\noindent{\textbf{Network Architecture.}}~~~~The decentralized actors and distributional reward estimation networks adopt the simple fully-connected feedforward neural network with three layers in our framework.
The two hidden layers' units are 64.
$|\mathcal{A}|$ and $2|\mathcal{A}|$ are the number of output units in actors and distributional reward estimation networks, respectively, where $|\mathcal{A}|$ represents the size of action space.
The centralized critic uses a graph attention neural network with eight attention heads, and each head's hidden unit is set to 8 to capture the dynamic relationship between agents.
After the graph attention neural network, a two-hidden-layer fully-connected feedforward neural network is used to get the state value. Besides, we choose \text{leaky relu} as nonlinear activation for all networks.

\noindent{\textbf{Compute.}}~~~~Experiments are carried out on NVIDIA GeForce RTX 1080Ti GPUs and with fixed hyperparameter settings, which are described in the following.
Each run of the experiments spanned about 2-12 hours, depending on the algorithm and the agents' number in the environment.

\noindent{\textbf{Hyperparameters.}}~~~~For all scenarios, the per-episode length is set to 25.
The periodical replay buffer learning rate is 0.4, the learning interval is set to 4 episodes, and the entropy scale is set to 0.3.
Detailed hyperparameters can be found in Table~\ref{DRE-MARL_hyper}.

\begin{table}[ht!]
\centering
\caption{The hyperparameters of DRE-MARL.}
\begin{tabular}{lc}
\toprule
Hyperparameter & Value\\
\midrule
optimizer & Adam~\cite{kingma2014adam} \\
learning rate & $1\cdot10^{-3}$ \\
entropy scale & 0.1/0.3 \\
\makecell[l]{number of hidden layers (all networks)} & 2 \\
\makecell[l]{nonlinearity of hidden layers (all networks)} & Leaky ReLU \\
\makecell[l]{number of hidden units per layer (all networks)} & 64 \\
number of attention head & 8 \\
time difference (TD) & 1 \\
buffer clear rate & 0.4 \\
discount ($\gamma$) & 0.95 \\
batch size & 1024 \\
$\varepsilon$-greedy & 0.7$\rightarrow$0.9 \\
tau ($\tau$) & 0.01 \\
$\alpha$ & 0.1 \\
$\beta$ & 10 \\
$\eta$ & 0.3 \\
\bottomrule
\end{tabular}
\label{DRE-MARL_hyper}
\end{table}

\begin{table*}[t!]
\centering
\caption{The comparison of the computational cost about different models based on the CN-3 scenario. The items of the comparison contain three aspects: parameters, physical training time (min), and max memory consumption (MB).}
\resizebox{\textwidth}{!}{
\begin{tabular}{c|l|c|c|c|c|c|c}
    \toprule
    \specialrule{0em}{1.5pt}{1.5pt}
    \toprule
    \multicolumn{2}{c|}{comparison items} & \multicolumn{6}{c}{model}\\
    \midrule[1pt]
     \multicolumn{2}{c|}{} & DRE-MARL & MAPPO & MAAC & QMIX & MADDPG & IQL\\
     \multirow{3}{*}{\rotatebox{90}{\makecell[c]{para-\\meters}}} &  total & 198440 & 72854 & 450364 & 307330 & 172884 & 564642\\
      & trainable & 54126 & 72854 & 450364 & 123488 & 86436 & 141147 \\
      & untrainable & 144314 & 0 & 0 & 183842 & 86448 & 423495\\
      \midrule[1pt]
      \multicolumn{2}{c|}{\makecell[c]{physical training\\ time (min)}} & 256.0\tiny{$\pm$1.859} & 336.0\tiny{$\pm$9.623} & 915.9\tiny{$\pm$6.270} & 867.1\tiny{$\pm$16.54} & 166.9\tiny{$\pm$1.976} & 217.4\tiny{$\pm$2.311}\\
      \midrule[1pt]
      \multicolumn{2}{c|}{\makecell[c]{max memory\\ consumption (MB)}} & 1354.\tiny{$\pm$60.18} & 310.6\tiny{$\pm$0.297} & 229590.1\tiny{$\pm$65.69} & 2146.5\tiny{$\pm$22.95} & 3754.\tiny{$\pm$227.3} & 668.4\tiny{$\pm$20.18} \\
    \bottomrule
    \specialrule{0em}{1.5pt}{1.5pt}
    \bottomrule
\end{tabular}
}
\end{table*}

\begin{table*}[t!]
\centering
\caption{Performance comparison of DRE-MARL with and without regularization term $L_R$ while training with the team rewards and evaluating without the $r_{ac\text{-}dist}$ setting. The values represent mean episodic rewards.}

\begin{tabular}{c|l|c}
    \toprule
    \specialrule{0em}{1.5pt}{1.5pt}
    \toprule
    \multicolumn{2}{c|}{comparison items} & \multicolumn{1}{c}{model}\\
    \midrule[1pt]
     \multicolumn{2}{c|}{} & DRE-MARL\\
     \multirow{2}{*}{$l_{SS}+g_{SS}$} &  with $L_R$ & -272.7\tiny{$\pm$25.22} \\
      & without $L_R$ & -274.6\tiny{$\pm$19.13}  \\
      \midrule[1pt]
      \multirow{2}{*}{$l_{SMO}+g_{SS}$} &  with $L_R$ & -235.1\tiny{$\pm$16.38} \\
      & without $L_R$ & -365.0\tiny{$\pm$25.19}  \\
      \midrule[1pt]
      \multirow{2}{*}{$l_{MO}+g_{MO}$} &  with $L_R$ & -242.1\tiny{$\pm$16.15} \\
      & without $L_R$ & -258.9\tiny{$\pm$17.36}  \\
      \midrule[1pt]
      \multirow{2}{*}{$l_{MO}+g_{SS}$} &  with $L_R$ & -252.6\tiny{$\pm$17.49} \\
      & without $L_R$ & -354.2\tiny{$\pm$23.47}  \\
      \midrule[1pt]
      \multicolumn{2}{c|}{no reward estimation} &  -258.2\tiny{$\pm$20.75}  \\
    \bottomrule
    \specialrule{0em}{1.5pt}{1.5pt}
    \bottomrule
\end{tabular}

\end{table*}

\begin{table*}[t!]
\centering
\caption{Performance comparison of DRE-MARL variants and DFAC variants. We conduct the experiment in CN-3 with DFAC variants which is based on SC II originally, and adopt the original hyperparameters of DFAC variants. The DFAC-diql(128) denotes the number of hidden neural is 128.}
\begin{tabular}{c|c}
    \toprule
    \specialrule{0em}{1.5pt}{1.5pt}
    \toprule
    \multicolumn{1}{c}{model} & performance\\
    \midrule[1pt]
     {$l_{SS}+g_{SS}$} & -272.7\tiny{$\pm$25.22} \\
    \midrule[1pt]
      {$l_{SMO}+g_{SS}$} &  -235.1\tiny{$\pm$16.38} \\
    \midrule[1pt]
     {$l_{MO}+g_{MO}$} & -242.1\tiny{$\pm$16.15} \\
    \midrule[1pt]
     {$l_{MO}+g_{SS}$} & -252.6\tiny{$\pm$17.49} \\
    \midrule[1pt]
     {DFAC-diql(128)} &  -786.9\tiny{$\pm$180.2}  \\
      \midrule[1pt]
     {DFAC-diql(256)} &  -1117.9\tiny{$\pm$21.38}  \\
      \midrule[1pt]
     {DFAC-dmix(128)} &  -1121.6\tiny{$\pm$21.51}  \\
    \bottomrule
    \specialrule{0em}{1.5pt}{1.5pt}
    \bottomrule
\end{tabular}

\end{table*}

\section{More discussion of limitations}

\noindent{\textbf{Continuous action space.}}~~~~In our method, the number of action branches matches the number of available discrete actions, so we can set the value of K to be equal to the number of available discrete actions such as “move forward”, “move backward”, “move left”, “move right”, and “motionless” in MPE. But in continuous action space, for example, we want to manipulate a robot arm. The “grasping” action needs us to assign a continuous value such as rotation angle to the robot arm. Right now, the available action is a range, so we can not define the number of K.
We discuss the following possible solutions to solve the problem of continuous action space.

\begin{itemize}
    \item One possible solution may be the discretization of the range of the available action value. More sophisticated discretization will bring better manipulation, but it will consume more computational resources at the same time. Although coarse discretization reduces the consumption of physical time, it may hurt the performance.
    \item Another possible solution is learning a network with actions as inputs, just like how we convert discrete-action DQN into continuous-action Q network in DDPG. The implicit reward function makes it possible to perform reward aggregation by sampling several discrete action points.
\end{itemize}

\noindent{\textbf{Prior Gaussian distribution.}}~~~~The reward distribution is indeed a bit hard to select in practice. However, just like tanglesome signals can be factorized into the superposition of simple but elementary sinusoidal signals, we may consider using a cluster of basic distributions as reward distribution in future work.

\end{document}